\documentclass{article}

\usepackage{microtype}
\usepackage{graphicx}
\usepackage{subcaption}
\usepackage{booktabs} % for professional tables
\usepackage{algorithmic}
\usepackage{algorithm}

\usepackage{hyperref}

\usepackage{arxiv}

\usepackage{hyperref}
\usepackage{enumitem}
\usepackage{amssymb} % for checkmark
\usepackage{amsmath}
\usepackage{amssymb}
\usepackage{mathtools}
\usepackage{amsthm}
\usepackage{multirow}
\usepackage{xcolor}
\usepackage{tcolorbox}
\usepackage{fontawesome5}
\tcbuselibrary{skins,breakable,hooks}

\hypersetup{
    colorlinks=true,
    urlcolor=blue,        % only affects \url and \href
    citecolor=black,      % keeps citations black
    linkcolor=black,      % keeps internal links (sections, figures) black
    filecolor=black
}

\usepackage[capitalize,noabbrev]{cleveref}

\theoremstyle{plain}

\theoremstyle{definition}

\theoremstyle{remark}

\usepackage[textsize=tiny]{todonotes}
\usepackage{natbib}

\title{Knowledge Graphs are Implicit Reward Models: Path-Derived Signals Enable Compositional Reasoning}

\renewcommand{\shorttitle}{KGs are Implicit Reward Models: Path-Derived Signals Enable Compositional Reasoning}

\author{
  Yuval Kansal \\
  Princeton University \\
  \texttt{yuvalkansal@princeton.edu}
  \And
  Niraj K. Jha \\
  Princeton University \\
  \texttt{jha@princeton.edu}
  }

\begin{document}
\maketitle

% Custom preprint notice at bottom of first page
\newcommand{\preprintnotice}[1]{%
  \def\@preprinttext{#1}%
  \let\@oldthanksfootnote\thanks%
  \renewcommand{\thanks}[1]{}%
  \footnotetext[0]{\@preprinttext}%
  \let\thanks\@oldthanksfootnote%
}

\blfootnote{Preprint. Under review.}




\begin{abstract}
  Large language models have achieved near-expert performance in structured reasoning domains like mathematics and programming, yet their ability to perform compositional multi-hop reasoning in specialized scientific fields remains limited. We propose 
  %that true human-level intelligence requires 
  a bottom-up learning paradigm in which models are grounded in axiomatic domain facts and compose them to solve complex, unseen tasks. To this end, we present a post-training pipeline, based on a combination of supervised fine-tuning and reinforcement learning (RL), in which knowledge graphs act as implicit reward models. By deriving novel reward signals from knowledge graph paths, we provide verifiable, scalable, and grounded supervision that encourages models to compose intermediate axioms rather than optimize only final answers during RL. We validate this approach in the medical domain, training a 14B model on short-hop reasoning paths (1-3 hops) and evaluating its zero-shot generalization to complex multi-hop queries (4-5 hops). Our experiments show that path-derived rewards act as a ``compositional bridge", enabling our model to significantly outperform much larger models and frontier systems like GPT-5.2 and Gemini 3 Pro, on the most difficult reasoning tasks. Furthermore, we demonstrate the robustness of our approach to adversarial perturbations against option-shuffling stress tests. This work suggests that grounding the reasoning process in structured knowledge is a scalable and efficient path toward intelligent reasoning. Our code is publicly available at: https://github.com/jha-lab/kg-implicit-reward-compositional-rl/.
  
\end{abstract}

\vspace{-0.5em}
\begin{center}
  \large\textbf{\faGlobe~Project Page:}~\href{https://kg-implicit-reward-model.github.io/}{\textcolor{blue}{\texttt{https://kg-implicit-reward-model.github.io/}}}
\end{center}

\section{Introduction}

Recent advances in language models have revealed that reasoning capabilities can be significantly enhanced through a combination of high-quality pretraining, supervised fine-tuning (SFT), carefully tuned reinforcement learning (RL)-based post-training, and strategic use of additional test-time compute \citep{OpenAIGPT52, GoogleDeepMindGemini3Pro, yang2025qwen3, muennighoff2025s1}. The resulting systems achieve near-expert performance in well-structured domains, such as mathematics and programming, where high-quality data have been curated, reasoning steps are clear, ground truth is unambiguous, and intermediate verification is tractable \citep{LightmanLetsStep, anthropic2025opus}. However, true human-level intelligence in specialized fields requires more than just general pattern matching or long-form generation; it requires \textbf{compositional reasoning}: the ability to reliably combine axiomatic facts for complex multi-hop problem solving \cite{kamp1995prototype, fodor1975language}. While current large language models (LLMs) excel when reasoning steps are clear and carefully curated expert data are available, compositional reasoning in high-stakes scientific domains, where reasoning paths are multi-faceted, remains elusive \citep{Yin2025TowardLearning, KimMedicalHealthcare}.

\begin{figure*}[ht]
  \vskip -0.5em
  \centering
  \includegraphics[width=1\textwidth]{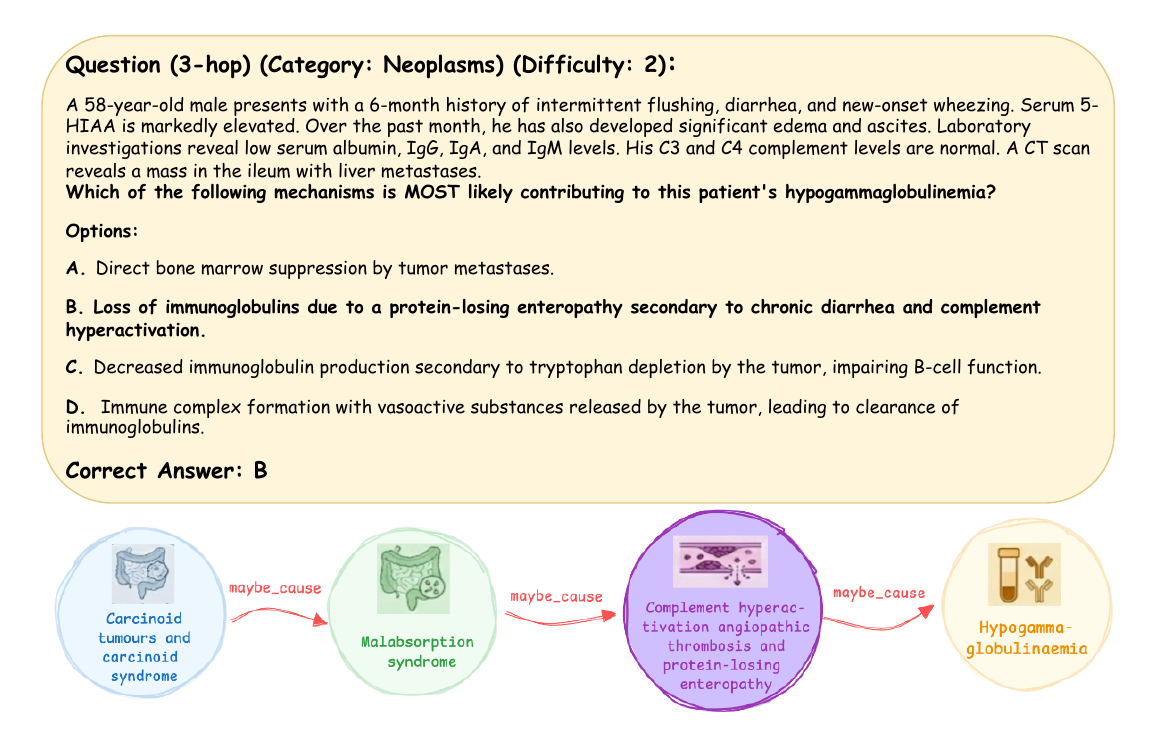}
  \caption{\textbf{Compositional Reasoning:} A sample 3-hop query that requires systematic traversal of axiomatic triples to make a grounded, multi-step clinical deduction.}
  \label{compositional-reasoning}
  \vskip -0.5em
\end{figure*}

To bridge this gap, we argue in favor of a \textbf{bottom-up learning paradigm}: grounding models in axiomatic facts and then composing these fundamentals into sophisticated domain knowledge. Knowledge graphs (KGs) provide a natural and promising scaffold for this grounding; they encode entities and relations in a structured, interpretable fashion that can represent the building blocks of domain knowledge at scale. Recent work  \cite{Dedhia2025Bottom-upNeed, Wang2024} has shown how high-quality data can be curated from such graphs and used to fine-tune models to obtain better reasoning traces. However, good static data is just the first step towards mastering the process of composition. Beyond high-quality data, robust reward design is a key lever for shaping models that can compose axiomatic facts from a domain to arrive at a logical conclusion. 

Existing post-training methods, e.g., reinforcement learning from human feedback \cite{ouyang2022training} and 
%its direct-optimization variants like 
direct preference optimization \cite{Rafailov2023DirectModel}, optimize models to match human preference with final outputs, not the process that produced them. Proxy reward signals, such as reward length and alignment with expert-written answers, while useful, fail to account for the composition intricacies needed to answer a complex multi-hop query. In practice, reward models often conflate superficial correlates (fluency, deference) with quality, thus leading to reward over-optimization and brittle answers \citep{Shrivastava2025SampleReasoning}. In safety-critical domains, the result is a mismatch between human-liked style and ground-truth validity \citep{DamaniBEYONDUNCERTAINTY, RewardLilLog, Rafailov2023DirectModel}. Whereas process supervision (rewarding intermediate steps) has shown promise in mathematics and logic \cite{Zhang2025TheReasoning, CuiPROCESSREWARDS, Wang2025HierarchicalModel, LightmanLetsStep}, curating and scaling expert-annotated data for other domains are notoriously difficult to achieve and nontrivial. This raises a key question: \textbf{How can we build systems and reward signals at scale that promote grounded compositional reasoning in multi-hop tasks without relying on expensive human-in-the-loop annotations?}

KGs offer an implicit solution to this scaling problem. In a KG, domain-specific concepts and their relationships are represented as axiomatic triples $(head, relation, tail)$. Our core insight is that by comparing the reasoning and assertions of a model during post-training against relevant triples and the chain of axiomatic facts required to solve the problem, we can turn the match (or mismatch) into a high-quality reward signal. Instead of an answer that ``looks good," this lets us reward the model to the degree its response is supported by verifiable domain knowledge and implicitly reward it for correctly composing facts to produce a solution. This is readily scalable without requiring external expert supervision and further enables us to move away from top-down distillation and ground the model’s reasoning in the field's fundamental building blocks.

In this article, we realize this idea through a Base Model $\rightarrow$ SFT [Low-Rank Adaptation (LoRA)]$\rightarrow$ RL [Group Relative Policy Optimization (GRPO)] post-training pipeline that uses a grounded KG to derive a novel reward signal to enable compositional reasoning \citep{hu2022lora, guo2025deepseek, yasunaga2021qa}. Whereas the approach can be generally applied, we study it in the medical domain, a field that serves as a rigorous stress test for compositional reasoning. Medical knowledge inherently requires multi-hop reasoning; a single clinical diagnosis may require navigating from a patient's demographics and medical history to symptoms, from those symptoms to a disease, and finally to a drug (a sample multi-hop query is shown in Fig.~\ref{compositional-reasoning}). By training a Qwen3 14B model \cite{yang2025qwen3} on simple 1-, 2-, and 3-hop reasoning paths derived from a KG, we probe whether it can learn the underlying ``logic of composition" to solve unseen, complex medical queries, ranging from 2- to 5-hop, in the ICD-Bench test suite \citep{Dedhia2025Bottom-upNeed}. Our results indicate that this grounded SFT+RL approach leads to large accuracy improvements on the most difficult questions, and remains robust under stress tests, such as option shuffling and ICD-10 category breakdowns \citep{whoICD10}. We find that while SFT provides the necessary knowledge base, RL acts as the ``compositional bridge." We demonstrate that insights learned on an 8B model transfer effectively to a 14B model, outperforming larger reasoning and frontier models. %TODO: mention some stats and cite Qwen

%TODO: see if need to make in para instead of list
% More concretely, we investigate these Research Questions:
% \begin{itemize}
%     \item \textbf{RQ1:} Can a grounded KG be used to automatically produce and scale reward functions for complex reasoning without human-in-the-loop?
%     \item \textbf{RQ2:} To what extent does training on short-hop compositional tasks enable zero-shot generalization to longer, more difficult questions?
%     \item \textbf{RQ3:} How does a KG-path-inspired reward compare to traditional outcome-based or distillation-based reward signals?
    
% \end{itemize}

Our core contributions can be summarized as follows:
\begin{itemize}
    \item \textbf{A Grounded, Scalable Reinforcement Learning with Verifiable Rewards (RLVR) Pipeline}: We introduce a scalable SFT+RL post-training framework designed to enable compositional reasoning in models using KGs as a verifiable ground truth.
    \item \textbf{KG-Path Inspired Reward}: We conduct a thorough investigation to design a novel reward signal derived from the KG that encourages compositional reasoning, correctness, and enables process supervision at scale.
    \item \textbf{Compositional Generalization}: We demonstrate how training on 1-to-3-hop paths enables a model to generalize to difficult and longer 4-, 5-hop questions, significantly outperforming base models and larger models.
    \item \textbf{Robustness \& Real-World Validation}: We stratify our model's performance by different difficulty levels, on real-world medical categories (ICD-10), and its resilience against adversarial option shuffling. 
\end{itemize}

% \textbf{A Grounded, Scalable Reinforcement Learning with Verifiable Rewards (RLVR) Pipeline}: We introduce a scalable SFT+RL post-training framework to enable compositional reasoning in models, using KGs as a verifiable ground truth. 

% \textbf{KG-Path Inspired Reward}: We conduct a thorough investigation to design a novel reward signal derived from the KG that encourages compositional reasoning and correctness, and enables process supervision at scale. 

% \textbf{Compositional Generalization}: We demonstrate how training on 1-to-3-hop paths enables a model to generalize to difficult and longer 4-, 5-hop questions, significantly outperforming base models and larger models. 

% \textbf{Robustness \& Real-World Validation}: We stratify the performance of our model by different difficulty levels, on real-world medical categories (ICD-10), and its resilience against adversarial option shuffling.

\section{Related Work}

%We discuss related work next.

\subsection{Role of SFT and RL in Reasoning}
Recent studies have intensely debated the distinct contributions of SFT and RL to model performance \citep{jin2025rl, kang2025quagmires, matsutani2025rl}. The authors of \cite{chu2025sft} argue that ``SFT memorizes, RL generalizes," claiming that while SFT stabilizes outputs, it struggles with out-of-distribution scenarios that RL can navigate. The authors of \cite{Rajani2025} characterize GRPO as a ``scalpel" that amplifies existing capabilities and SFT as a ``hammer" to overwrite prior knowledge. Our findings align with these dynamics. We use SFT to instill atomic domain knowledge in the model and RL to amplify compositional logic required to connect such knowledge.

% Our findings contrast with skepticism in \cite{yue2025does}, who suggest that current RL methods fail to incentivize reasoning capacities beyond the base model's potential unless specific conditions are met. Our results on unseen 4-, 5-hop queries demonstrate that when rewards are grounded in relevant axiomatic primitives, RL can elicit novel compositional abilities beyond the baseline.

Contrary to the findings in \cite{yue2025does}, our results on unseen 4-, 5-hop queries demonstrate that when rewards are grounded in relevant axiomatic primitives, RL can elicit novel compositional abilities beyond the baseline. This echoes the findings in \cite{yuan2025f} that demonstrate that RL can teach models to compose old skills into new ones; we validate this in a high-stakes real-world domain rather than a synthetic one. 

% \enlargethispage{2\baselineskip}
\subsection{RL on KGs}
Traditional applications of RL on KGs, e.g., \cite{DasGOLEARNING} and \cite{xiong2017deeppath}, primarily focus on traversing graph structures to complete missing triples (link prediction) or find missing entities. The authors of \cite{lin2018multi} further refine this approach with reward shaping to improve multi-hop reasoning, but still largely confine themselves to the task of graph completion instead of open-ended question answering in a real-world setting. In more recent works, the authors of \cite{Wang2024} propose ``Learning to Plan," where KGs guide the retrieval process for retrieval-augmented generation systems, and those of \cite{YanRLKGF} introduce RL from KG feedback to replace human feedback with KG signals. While promising, these approaches often limit the role of the KG to retrieval planning or simple search tools for alignment.

Our work differs fundamentally by centrally positioning KGs as a dense process verifier for real-world multi-hop reasoning. The authors of \cite{khatwani2025brittleness} use LLMs as a reward model for KG reasoning, but found the approach to be brittle, with poor transfer to downstream diagnostic tasks. We attribute this to the lack of a compositional training curriculum; by combining the bottom-up data curation of \cite{Dedhia2025Bottom-upNeed} with treating the KG as a reward model to derive path-aligned signals, we overcome this limitation. Furthermore, unlike \cite{GunjalRubrics} that uses unstructured LLM-created rubrics as rewards, or rule-based Logic-RL \cite{XieLogicRL}, we derive our signal directly from grounded axiomatic paths of the KG. 

\section{Preliminaries}

%We provide some background material next.

\subsection{Notation and RL for Language Models}
We treat an LLM as a stochastic policy $\pi_\theta$ that maps a query $q$ [multiple-choice question (MCQ) task] to a distribution over possible completions $y$. Each completion $y$ comprises a reasoning trace $r$ (chain-of-thought) \cite{wei2022chain} and a final answer $\hat{a}$ (spanning A-D). Each training task $q$ is associated with a ground-truth answer $a^*$ and a ground-truth KG path $P={(h_i,r_i,t_i)}^L_{i=1}$ (see Section \ref{Data_Construction}). 

A composite scalar reward function $R(y)$, derived from the KG, is used to score a generated completion $y$ (see Section \ref{Reward_func}). The RL objective is to maximize the expected reward under the prompt distribution: 
% \vspace*{-1mm}
\[
\mathbf{J}(\theta) = \mathbf{E}_{q\sim D}\mathbf{E}_{y\sim\pi_{\theta}(\cdot|q)}[R(y)]
%\vspace{-0.5em}
\]
Whereas the response $y$ is produced token-by-token, we treat the entire completion as a single trajectory for reward assignment, following common practice in LLM post-training.

Policy updates are performed using GRPO, a popular proximal policy optimization-like optimizer \cite{schulman2017proximal} that drops the critic and estimates advantages at the group level using normalization \citep{guo2025deepseek}. See Appendix \ref{Appendix E} for details of our hyperparameter configuration for the SFT and RL stages.

% \enlargethispage{2\baselineskip}
\subsection{SFT Followed by RL} %TODO: add some citations
Our training recipe follows the widely adopted Base Model $\rightarrow$ SFT $\rightarrow$ RL framework for improving LLMs. First, an SFT stage initializes the policy (base LLM) to produce high-quality, KG-grounded reasoning traces. A subsequent RL stage refines the policy directly to optimize the reward signal and enable compositional reasoning. Formally, SFT minimizes the negative log-likelihood on a supervised dataset consisting of question-answer (QA) tasks paired with reference reasoning traces and answers.

In our experiments, the SFT stage provides broad KG coverage, whereas the RL stage is deliberately small: a design choice motivated by observed instability in an RL-from-scratch approach and because targeted RL with good reward can enable compositional abilities when built atop SFT initialization (Section \ref{RL_ablations}, Appendix \ref{Appendix G} have more details).

\subsection{Medical KG: Unified Medical Language System (UMLS)}
We instantiate our framework on a standard biomedical KG based on UMLS \cite{bodenreider2004unified}, which encodes canonical medical ontology in a structured graph format. Each fact is represented as a triple $(head, relation, tail)$ and multi-hop paths $P = {(h_i, r_i, t_i)}_{i=1}^L$ serve as axiomatic compositional primitives used to generate QA tasks and path-alignment reward signals, and evaluate correctness. Full QA task generation and reward-design choices/strategies are described in the following section.
%Section \ref{Methodology}.

% \enlargethispage{2\baselineskip}
\section{Methodology}
\label{Methodology}
\begin{figure*}[ht]
  \vskip 0.2in
  \centering
  \includegraphics[width=1\textwidth]{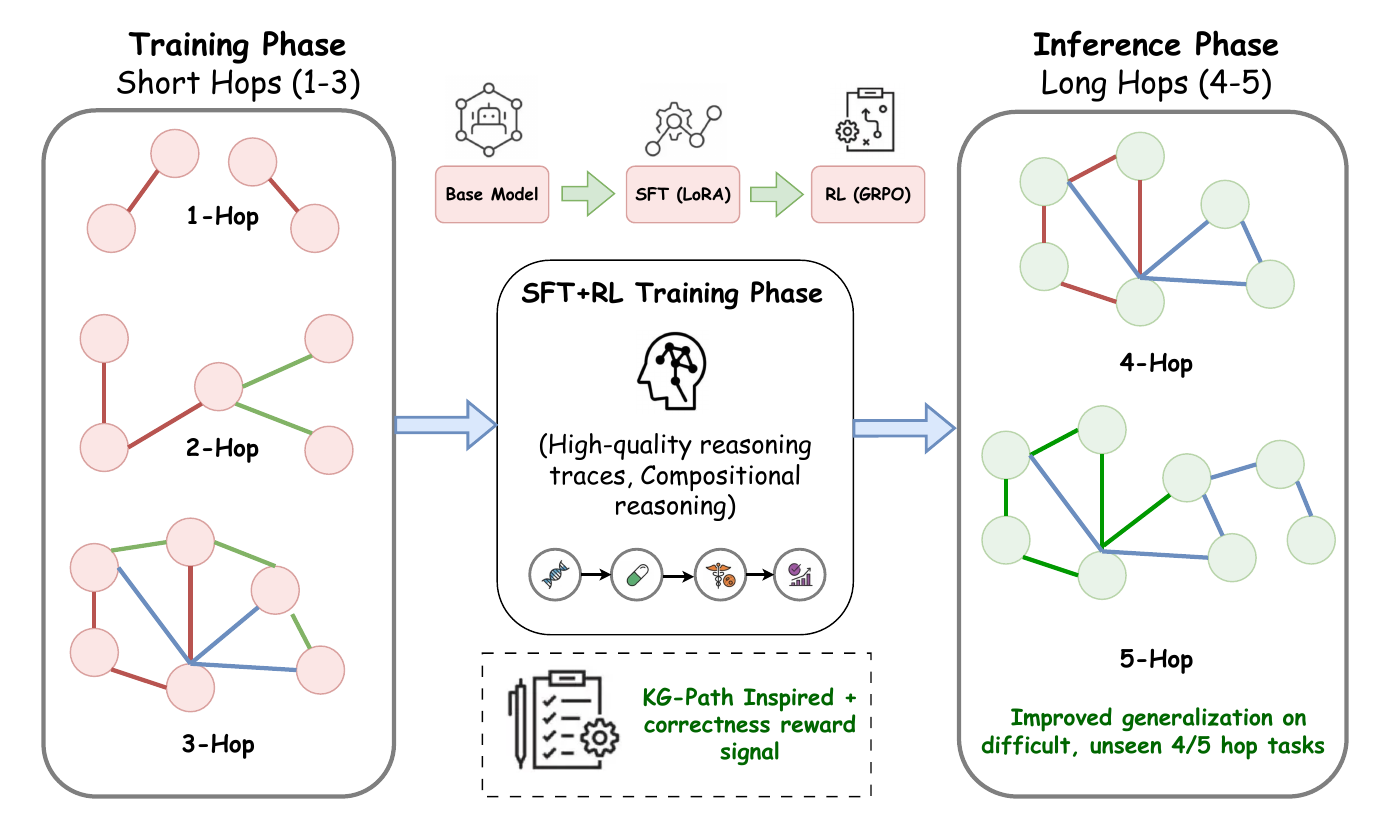}
  \caption{
    \textbf{SFT+RL pipeline overview:} Schematic of the pipeline from SFT to KG-grounded RL. While SFT enables domain-specific grounding, the path-derived reward signal during RL provides the process supervision necessary for compositional reasoning.
  }
  \label{overall_pipeline}
\end{figure*}

This section describes our data construction and training pipeline. We emphasize the sequential decisions we make: dataset choices, SFT warm start, RL budget, reward design, and experiments that guide these choices. Our training pipeline is designed to transition a base model from broad competence to deep, compositional medical-domain reasoning. Fig.~\ref{overall_pipeline} presents an overview of our pipeline. %TODO: more details in the appendix

\subsection{Data Construction and Axiomatic Grounding}
\label{Data_Construction}
Given our testbed that involves compositional reasoning and to ensure our model learns true depth rather than mere pattern matching, we adopt the data-generation and curation pipeline from \cite{Dedhia2025Bottom-upNeed}, which enables scalable generation of multi-hop reasoning questions grounded in a medical KG. The structured representation of medical concepts, such as diseases, drugs, signs, and mechanisms, in a KG as $(head, relation, tail)$ triples enables question generation directly from verifiable and grounded KG paths. Questions are generated in natural language in an MCQ format using a backend LLM by traversing $n$-length paths within the KG, where $n$ represents the number of ``hops" required to link a starting node to a final node. This enables precise control over the compositional complexity of each query. Furthermore, each question is paired with a rich reasoning trace and a ground truth path: a sequence of $(head, relation, tail)$ triples that constitutes a verifiable logical chain. The pipeline helps stratify questions by hop length, difficulty, and ICD-10 category, and enforces strict separation between the training and test sets at the path and entity levels to avoid leakage. 

We generate a training set of 24,660 QA tasks designed to ensure maximum node coverage across the KG \citep{yasunaga2021qa}. For evaluation, we use ICD-Bench, a non-overlapping test set of 3,675 questions \citep{Dedhia2025Bottom-upNeed}. Importantly, the training set consists of 1-3 hop paths, whereas ICD-Bench includes 2-5 hop path tasks across 15 ICD-10 categories to test zero-shot compositional generalization at different difficulty levels [1 (very easy) - 5 (very hard)]. More analysis of overlaps between the training and test sets is presented in Appendix \ref{Appendix D}.

Our training pipeline consists of three stages: Base Model $\rightarrow$ SFT (LoRA) $\rightarrow$ RL (GRPO). This design reflects our central hypothesis that compositional reasoning emerges most reliably when models are first grounded in rich reasoning traces via supervised learning and then tuned using scalable process-aligned rewards derived from the KG paths. We emphasize that all training data and rewards are derived from the same KG to ensure consistency between training and evaluation. 

% \enlargethispage{2\baselineskip}
\subsection{RL Alone is Insufficient}
\label{RL_ablations}
A core finding of our work is that the Zero-RL approach, applying GRPO directly to the base LLM, is insufficient for deep domain expertise at our model scale. The model requires an understanding of the domain axioms before it can learn to compose. Starting from the base model (Qwen3 8B), we apply GRPO directly using subsets of training data: 5k, 10k, and all 24.66k examples. Across these settings, we find that while RL improves performance relative to the base model, it does not consistently outperform SFT-only training on the base model. Interestingly, the budget of 5k examples yields the strongest results among all other settings, suggesting how large-scale vanilla RL without proper grounding is insufficient for compositional behavior, motivating the use of SFT for an initial warm start. 

Based on these observations, we use 5k examples in the RL stage and the remaining 19.66k in the SFT stage. The base model is fine-tuned using LoRA on 19.66k examples, followed by GRPO on the remaining 5k examples as per the SFT+RL pipeline.

See Appendix \ref{Appendix A}, \ref{Appendix B} for detailed ablations with the Zero-RL and SFT+RL pipelines on the Qwen3 8B model, respectively. See Appendix \ref{Appendix H} for our GRPO training prompt.

% \enlargethispage{2\baselineskip} % TODO: check if this okay
\subsection{Reward Design Exploration for Compositional Reasoning} %TODO: put more details in the appendix on the structure of rewards
A central goal and a novel contribution of this work is to design scalable reward signals during training that enable compositional reasoning beyond surface-level understanding. Towards this end, we conduct extensive ablation studies using a combination of four distinct reward signals to determine which best fosters verifiable composition:

\begin{itemize}
    \item \textbf{Binary Correctness $(R_{bin})$:} A simple outcome-based signal that rewards the final answer. 
    \item \textbf{Similarity $(R_{sim})$:} A distillation-based reward that measures the Jaccard similarity between the model output and an expert reasoning trace (generated by Gemini 2.5 Pro during data curation).
    \item \textbf{Thinking Quality $(R_{think})$:} A reward designed to score the thinking quality and length of the generation.
    \item \textbf{Path Alignment $(R_{path})$:} A novel reward that scores the model based on the coverage of the ground-truth KG triples in the model response.
\end{itemize}

% \textbf{Binary Correctness $(R_{bin})$:} A simple outcome-based signal that rewards the final answer. 

% \textbf{Similarity $(R_{sim})$:} A distillation-based reward that measures the Jaccard similarity between the model output and an expert reasoning trace (generated by Gemini 2.5 Pro during data curation).

% \textbf{Thinking Quality $(R_{think})$:} A reward designed to score the thinking quality and length of the generation.

% \textbf{Path Alignment $(R_{path})$:} A novel reward that scores the model based on the coverage of the ground-truth KG triples in the model response.

We conduct a systematic exploration by evaluating a combination of these rewards, always including $R_{bin}$ (+1 for correctness, 0 otherwise) as a minimal signal. Empirically, we discover that $R_{think}$ is often unstable and leads to reward hacking, generating inefficacious chains. $R_{sim}$ also proved sub-optimal, suggesting distillation rewards over-optimize aesthetic mimicry rather than true logical composition. 

Our findings highlight the power of simplicity: The combination of path alignment and binary correctness provides the strongest signal for composition. Whereas $R_{bin}$ optimizes correctness, $R_{path}$ rewards the model for identifying and applying the axiomatic facts (triples) required to compose the correct solution. To further strengthen the outcome signal, we replace the simple $R_{bin}$ with negative sampling reinforcement \cite{Zhu2025}, which penalizes incorrect generations by upweighting the negative reward, thereby encouraging the model to explore alternative/correct trajectories.

\subsection{KG-Grounded Reward Formulation}
\label{Reward_func}
To provide a robust reward signal to enable composition, we develop a composite reward that balances outcome correctness with path-level alignment grounded in the KG. Let the model generate a response $y$ for a question $q$ with the reasoning trace $r$ and a final answer $\hat{a}$. For each QA task, there is a ground truth answer $a^*$ and a ground-truth KG path $P = {(h_i, r_i, t_i)}_{i=1}^L$, where $L$ is the path length. The total reward is then a combination of the two rewards:
\[
\vspace{-0.5em}
R_{total}(y) = R_{bin}(\hat{a}, a^*) + R_{path}(r,P)
\vspace{-0.5em}
\]

\textbf{Binary Correctness Reward:} This provides a minimal but necessary supervision signal on the final answer.
\[
\vspace{-0.5em}
R_{\text{bin}}(\hat{a}, a^*) =
\begin{cases}
\alpha, & \text{if } \hat{a} = a^* \\
-\beta, & \text{otherwise}
\end{cases}
\vspace{-0.5em}
\]

where $\alpha, \beta > 0$ and $\beta > \alpha$. This asymmetric design ensures stable learning by reinforcing exploration of correct alternate paths. We use $\alpha = 0.1, \beta = 1$ in accordance with investigations by \citep{Zhu2025}.

\textbf{Path Alignment Reward (KG-grounded):} The primary technical innovation is $R_{path}$, which provides automated and scalable process supervision by evaluating whether the reasoning trace of the model aligns with the ground-truth KG path $P$ during RL post-training. We first tokenize and normalize the reasoning trace $r$ to extract a set of textual tokens, $T(r)$. We derive a corresponding set of path tokens, $T(P)$, from the ground-truth path $P$, representing the entities in $P$. The core signal is path coverage: 
\[
\vspace{-0.5em}
\text{coverage}(r, P) = \frac{\mid T(r)\cap T(P)\mid}{\mid T(P)\mid}
\vspace{-0.5em}
\]

We include a minimum-hit constraint that requires alignment with at least two distinct path entities to discourage trivial matches and promote logical composition. We apply an additional repetition penalty to reduce reward hacking and avoid linguistic collapse. The final reward is defined as:
\[
\vspace{-0.5em}
\begin{split}
R_{path}(r, P) = &\min(\gamma_1\cdot \text{coverage}(r, P) \\
&+ \gamma_2\cdot \mathbf{I}(|T(r)\cap T(P)|\geq 2), R_{max}),
\end{split}
\vspace{-0.5em}
\]

scaled by a repetition penalty factor, $\phi_{rep}$, and clipped to a fixed maximum. We use $\gamma_1 = 1.2, \gamma_2 = 0.3$, $R_{max} = 1.5$.

The total reward, $R_{total}$, is process-level, grounded, and compositional. It is readily scalable and easily verifiable as a result of the grounding in the KG. Furthermore, unlike similarity-based distillation or AI-based rubric rewards \cite{GunjalRubrics}, alignment is scored against true domain structure rather than stylistic mimicry. We present a formal formulation of $R_{sim}$ and $R_{think}$ in Appendix \ref{Appendix C}.

\subsection{Scaling and Benchmarking}
We evaluate our pipeline on the Qwen3 8B model before scaling the findings to the 14B variant without modification. The 14B model trained on our pipeline not only generalizes to 2- and 3-hop tasks, but also to 4-and 5-hop tasks with remarkable efficacy, surpassing much larger frontier models. In addition to overall accuracy, we stratify performance by hop length, difficulty level, ICD-10 category, and robustness under the option-shuffling stress test. 

\section{Results}
\textbf{Setup:} We initially evaluate three systems: Base Qwen3 14B model, model trained using LoRA on the full training set (24,660 QA tasks), and our proposed SFT+RL pipeline (SFT on 19,660 tasks followed by GRPO on the remaining 5k tasks) based on leveraging the KG-derived reward. We do not report results for models trained using the Zero-RL approach, described in Section \ref{RL_ablations}, since all of them performed worse than SFT on the full dataset (see Appendix \ref{Appendix A} for details). All evaluations use the held-out test ICD-Bench test set (3,675 tasks) to validate whether path-derived signals truly enable compositional reasoning. Finally, we evaluate our SFT+RL model against larger frontier and reasoning models. See Appendix \ref{Appendix F} for sample model responses of our final Qwen3 14B SFT+RL model.

\subsection{Scaling Composition: From Short-Hop Training to Long-Hop Reasoning}
The primary claim of this work is that KGs function as implicit reward models, and grounding the model reasoning in path-derived signals enables it to learn the underlying logic of composition rather than regurgitating information seen during training.

\begin{figure}[h!]
  \vskip -0.5em  % negative space instead of positive
  \begin{center}
    \centerline{\includegraphics[width=0.8\textwidth]{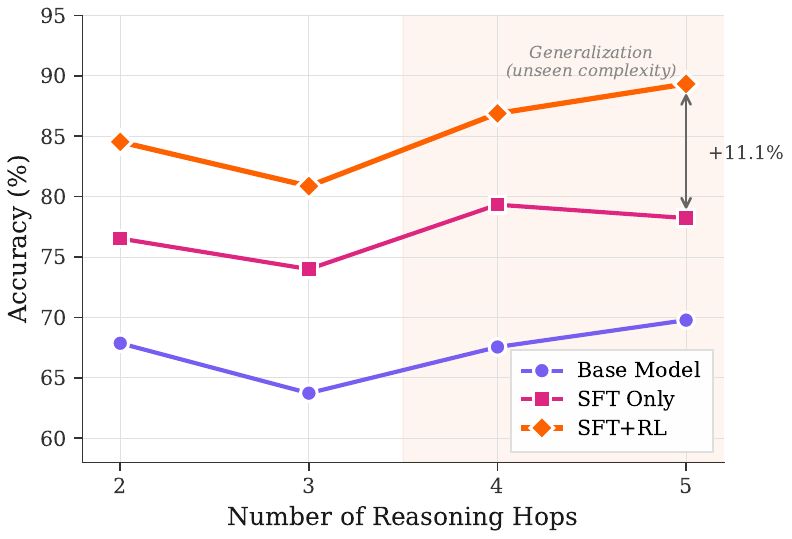}}
    \caption{
      \textbf{Accuracy by Hop Length:} Our SFT+RL model not only outperforms baselines on 2-3 hop tasks but exhibits a positive performance gradient on unseen 4-, 5-hop reasoning tasks, validating the ``compositional bridge" enabled by path-aligned rewards.
    }
    \label{hop_length_acc}
  \end{center}
  \vskip -0.5em  % add negative space after too if needed
\end{figure}
\vspace{-1em}

\textbf{Path-derived Signals Enable Compositional Reasoning:} Whereas the model was exposed to 1-, 2-, and 3-hop paths during the SFT+RL training phase, it remained totally naive to tasks involving 4-, 5-hop reasoning. As shown in Fig.~\ref{hop_length_acc}, the SFT+RL model demonstrates substantially stronger generalization to longer paths, achieving a notable gain of 7.5\% on unseen 4-hop and 11.1\% on unseen 5-hop questions relative to the SFT-only approach. This improvement is not attributable to exposure to longer chains, since the training and evaluation distributions are identical across all models. Instead, it reflects the effect of path-derived signals introduced during RL. By rewarding assertions that align the model directly with the ground-truth KG path, the model learns the logic of composition. Importantly, the generalization gap between the SFT-only and SFT+RL approaches widens as hop-length increases. This is a hallmark of genuine compositional learning. The KG-derived reward signal $(R_{path})$ enables the model to decompose long-horizon reasoning into verifiable steps and compose reasoning beyond the complexity observed during training. 

\subsection{Robustness to Tasks Involving Reasoning Depth}
A significant challenge in medical reasoning is maintaining integrity as the complexity of tasks increases. Aggregate performance metrics often obscure model failure on long-tail high-difficulty scenarios, where deep reasoning is paramount. We assign difficulty ratings from 1 (very easy) to 5 (very hard) \cite{Dedhia2025Bottom-upNeed} to each ICD-Bench question and assess model performance as a function of question difficulty.

\begin{figure}[h!]
  \vskip -0.5em  % negative space instead of positive
  \begin{center}
    \centerline{\includegraphics[width=0.8\columnwidth]{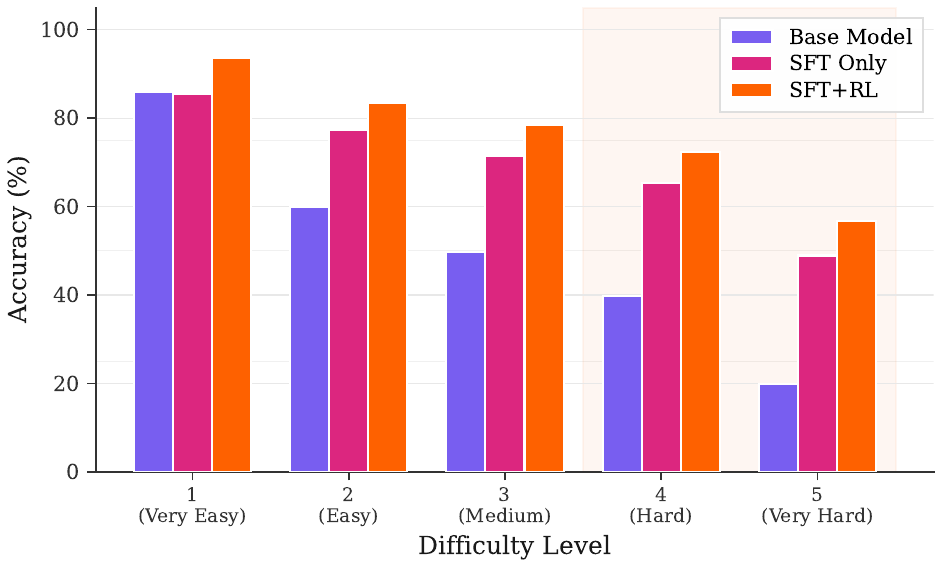}}
    \caption{
      \textbf{Accuracy by Difficulty Level:} Whereas the Base Model’s reasoning collapses as task complexity increases, the SFT+RL pipeline exhibits robustness, maintaining a consistent lead over the SFT-only baseline across all levels.
    }
    \label{difficulty_acc}
  \end{center}
  \vskip -0.5em  % add negative space after too if needed
\end{figure}
\vspace{-1em}

\textbf{Dominance in High-Complexity Tasks:} The results shown in Fig.~\ref{difficulty_acc} demonstrate that the SFT+RL pipeline, grounded in path-derived signals, provides the most significant gains as task complexity increases. On Level-5 tasks, the base model accuracy collapses to 19.94\%, indicating worse-than-random-guess accuracy in complex clinical scenarios. Whereas the SFT-only approach improves this to 48.93\%, our SFT+RL model achieves 56.75\%, nearly tripling the base model performance. %TODO: say how this is better than QwQ-Med-3, also and even 5.2, Gemini 3. 

\textbf{Consistency across all Difficulty Levels:} On Level-1 tasks, our model reaches a near-ceiling accuracy (93.49\%) and the performance gap remains robust across difficulty levels, with the SFT+RL model consistently maintaining a 7-10\% lead over the SFT-only model. This demonstrates that by using the KG as an implicit reward model, we raise the performance floor for the most complex queries. On Level-5 tasks, where even large frontier models struggle, maintaining over 56\% accuracy is a testament to how path-aligned rewards do more than just stylistic inference; they enable the model to reliably compose multi-step chains. %TODO: Add heatmap in appendix

\begin{figure}[h!]
  \vskip -0.5em  % negative space instead of positive
  \begin{center}
    \centerline{\includegraphics[width=0.8\columnwidth]{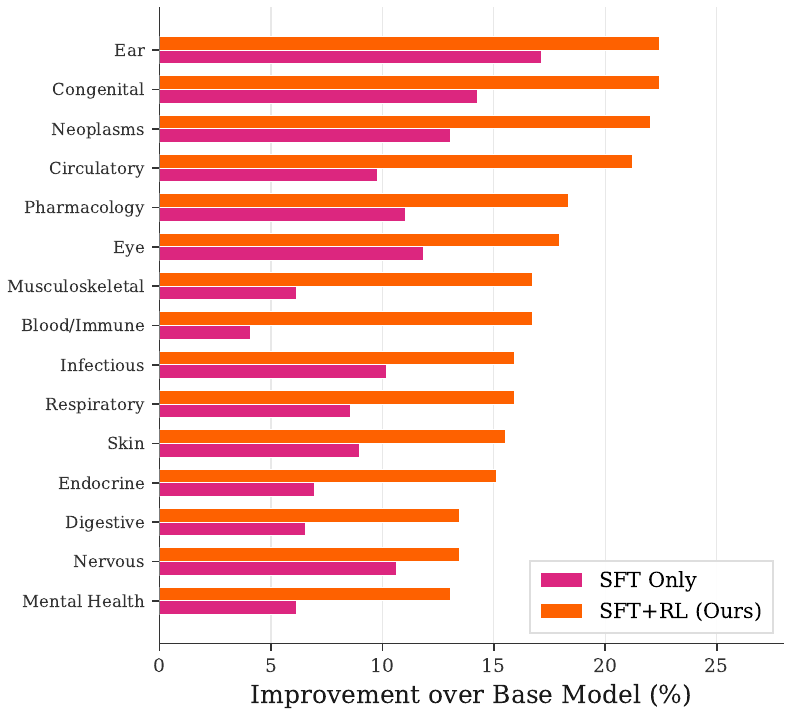}}
    \caption{
      \textbf{Accuracy by ICD-10 Category:} Path-aligned rewards consistently improve performance across all 15 medical sub-domains.
    }
    \label{icd_10_acc}
  \end{center}
  \vskip -1em  % add negative space after too if needed
\end{figure}
\vspace{-1em}

% \enlargethispage{2\baselineskip}
\subsection{ICD-10 Category Analysis: KG-grounded Gains Are Broadly Distributed}
To connect our empirical evaluation to real-world, clinically meaningful structures, we follow the categorization introduced in \cite{Dedhia2025Bottom-upNeed} that groups ICD-Bench questions into 15 ICD-10 categories in the data curation step. This classification lets us evaluate the breadth of our approach, enabling us to measure whether improvements of the SFT+RL model are concentrated in a few medical sub-domains or distributed across practical specialties.

The analysis, visualized in Fig.~\ref{icd_10_acc}, reveals that our SFT+RL pipeline consistently achieves the highest accuracy and maintains a substantial lead over the SFT-only baseline across all medical categories. Some of the most significant improvements occur in high-stakes areas such as ``Blood and Immune System Diseases" and ``Circulatory System Diseases," where diagnostic reasoning often requires complex, multi-hop evidence composition. This is consistent with our design choice to derive the reward from KG paths; the path-alignment reward can assign a meaningful intermediate signal and better shape compositional behavior.

\subsection{Robustness to Format Perturbation}
A common failure mode for LLMs is reliance on superficial cues, e.g., the order of options and answer in a multiple-choice list, rather than the true logical content and chain of reasoning. To evaluate the robustness of our models against such positional bias, we subject them to Stress-Test 3 (option shuffling) %as defined in recent work 
\cite{gu2025illusion}. In this test, the order of incorrect distractor options is randomized while keeping the correct answer choice constant.

\begin{table}[ht]
\caption{Analysis of Option Format Perturbation.}
\label{tab:shuffled_evidence}
\vskip 0.5em
\begin{center}
\begin{small}
\begin{sc}
\begin{tabular}{lccc}
\toprule
Method & Standard & Shuffled & Delta $(\Delta)$ \\
\midrule
SFT-Only & 75.95\% & 74.91\% & $-1.04\%$ \\
SFT+RL (Ours) & \textbf{83.62\%} & \textbf{82.45\%} & $-1.17\%$ \\
\bottomrule
\end{tabular}
\end{sc}
\end{small}
\end{center}
\vskip -0.5em
\end{table}

% \enlargethispage{2\baselineskip}
The results in Table \ref{tab:shuffled_evidence} demonstrate a remarkable degree of robustness of our approach compared to frontier models analyzed in the literature. Whereas leading systems, such as GPT-5 and Gemini-2.5 Pro, have been shown to suffer performance drops of 4-6\% under similar perturbations (in text-only medical contexts) \cite{gu2025illusion}, our models maintain nearly stable performance with a negligible drop of $\sim1\%$. 

\textbf{Importance of High-Quality Data:} This resilience highlights a fundamental axiom of machine learning: use of high-quality, grounded data is paramount. Our data curation and training pipeline incentivizes the model to identify the correct answer based on a verifiable reasoning path rather than shortcut patterns seen during unstructured training. The fact that even the SFT-only model trained on high-quality traces exhibits such stability suggests that grounding the model in structured domain axiomatic knowledge is as critical as subsequent RL optimization. By grounding the model and ensuring it learns true logical composition, we move closer to systems capable of genuine domain competence rather than ``illusion of readiness."

\subsection{Algorithmic Efficiency vs.~Scale: Surpassing Frontier Models}
One of our central claims is that careful reward design and bottom-up data curation can enable compositional reasoning that outperforms top-down brute-force scaling. To validate this, we compare our 14B SFT+RL model against two distinct classes of benchmarks: (1) large frontier models (GPT-5.2, Gemini 3 Pro) that represent the ceiling of generalist zero-shot reasoning, and (2) QwQ-Med-3 (32B), a domain-expert model distilled for medical reasoning.

\begin{figure}[h!]
  \vskip -0.5em  % negative space instead of positive
  \begin{center}
    \centerline{\includegraphics[width=0.8\columnwidth]{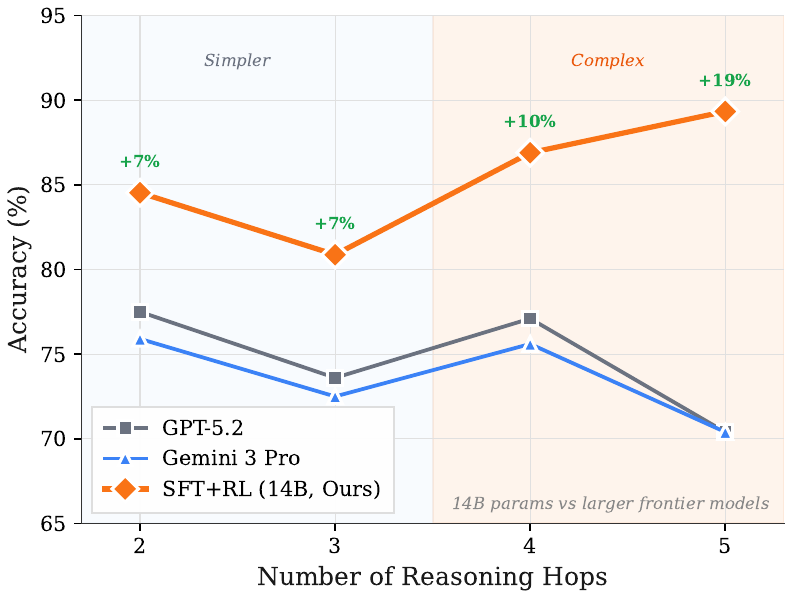}}
    \caption{
      \textbf{Accuracy Comparisons against Frontier Models by Hop Level:} Whereas the accuracy of generalist giants decays on longer chains, our 14B SFT+RL model achieves its highest accuracy on unseen 5-hop queries, validating the KG as a superior supervisor for complex composition.
    }
    \label{frontier_comparison}
  \end{center}
  \vskip -0.5em  % add negative space after too if needed
\end{figure}
% \vspace{-1em}

% \vspace*{-2mm}
\textbf{Superiority over Frontier Models:} We evaluate the zero-shot performance of the model against leading generalist models in Fig.~\ref{frontier_comparison}. Our results demonstrate that grounding a smaller model in a domain's axioms enables it to surpass larger, generally trained giants at complex reasoning tasks. We also note a striking trend: Whereas GPT-5.2 and Gemini 3 Pro maintain respectable performance on shorter hops, their accuracy stagnates or declines as the hop count increases. In contrast, our model exhibits a positive compositional gradient and achieves its highest accuracy (89.33\%) on the 5-hop queries. This supports our claim that KG-grounded path-derived signals teach the model how to compose axioms, not just standard pattern matching.

\textbf{Surpassing Expert-distilled Scale (SFT-Only)}: Finally, we compare our model with the QwQ-Med-3 model presented in \cite{Dedhia2025Bottom-upNeed}, which was trained on a similar data distribution. For a fair comparison, we use the majority-voting ($n=16$) metric, matching the aggregation metric used in their study (see Table \ref{tab:difficulty}).

\begin{table}[ht]
\caption{Performance by difficulty level (using majority voting).}
\label{tab:difficulty}
\vskip 0.15in
\begin{center}
\begin{small}
\begin{sc}
\begin{tabular}{ccccc}
\toprule
Diff. & QwQ-Med-3 32B & Ours-14B & $\Delta$ \\
\midrule
1 & \textbf{96.75\%} & 94.23\% & $-2.52\%$ \\
2 & 83.79\% & \textbf{85.63\%} & $+1.84\%$ \\
3 & 79.33\% & \textbf{80.33\%} & $+1.00\%$ \\
4 & 70.56\% & \textbf{71.50\%} & $+0.94\%$ \\
5 & 49.69\% & \textbf{59.05\%} & $+9.36\%$ \\
\bottomrule
\end{tabular}
\end{sc}
\end{small}
\end{center}
% \vskip -0.1in
\end{table}

Whereas the QwQ-Med-3 model holds a slight advantage on lower-difficulty tasks, which typically rely on factual recall, our model successfully bridges the recall-reasoning gap. By using the KG as an implicit reward model, we enable a smaller architecture to ``out-reason" a much larger model. This confirms our premise that while scale is a powerful tool for breadth of knowledge, path-aligned rewards are the true bridge to deep compositional reasoning.

\section{Discussion}

The ideas and results presented in this work highlight a fundamental shift in how we approach the development of expert-level reasoning. By using a KG as an implicit reward model, we demonstrate how bottom-up primitives in a domain can serve as highly scalable, verifiable process supervisors. Unlike contemporary human-in-the-loop process supervision, which is prohibitively expensive and infeasible to scale to millions of reasoning chains across domains, our path-derived reward signal provides an automated, scalable axiomatic grounding mechanism. 

Furthermore, our findings reinforce the basic axiom of machine learning: \textbf{Good data are paramount!} We show that when models are trained with proper axiomatic grounding, they develop a robust skill for compositional reasoning beyond simple factual recall. This is most evident by our model's ability to surpass much larger generalist giants. While brute-force scaling continues to dominate the search for general intelligence, our work suggests a more efficient path towards building superintelligent systems: building small, specialized models that master composition within their respective domains.

Finally, our reward design mechanism is inherently scalable and domain-agnostic. Any scientific or technical field that can be represented as a structured KG (from organic chemistry to case law) is a candidate for this pipeline. As domain KGs continue to expand in coverage and fidelity, they offer a practical route to building systems that reason from first principles rather than surface correlations or simple pattern matching. We posit this work as an early step toward scalable, verifiable domain-specific superintelligence, and we encourage future research to explore richer graph structures, broader domains, and tighter integration between symbolic knowledge and neural architectures to build better reasoning systems.

\section{Conclusion}
We introduced a simple, general idea: Treat KGs as implicit reward models and use path-derived rewards in a scalable fashion to teach models how to compose domain primitives into longer reasoning chains. We combined SFT with a compact RL stage and a KG path-aligned reward (plus correctness) to synthesize models that can generalize from 1-3-hop training to unseen 4-5-hop problems, improve baseline accuracy on the hardest tasks, and remain robust under format perturbations. Our recipe yields consistent gains over an SFT-only approach and can outperform much larger models, validating that good grounded data and reward design, not just scale, are central to compositional reasoning. Although we have successfully validated our claims within the medical domain, we acknowledge that this is only a starting point. We encourage the community to further investigate how structured KGs can be leveraged as reward supervisors to build the next generation of superintelligent systems.

\section*{Acknowledgments}

The experiments reported in this paper were performed on the computational resources managed and supported by Princeton Research Computing, Princeton AI Lab, and the Princeton Language and Intelligence Initiative at Princeton University.


\newpage
\bibliography{example_paper}

\begin{thebibliography}{43}
\providecommand{\natexlab}[1]{#1}
\providecommand{\url}[1]{\texttt{#1}}
\expandafter\ifx\csname urlstyle\endcsname\relax
  \providecommand{\doi}[1]{doi: #1}\else
  \providecommand{\doi}{doi: \begingroup \urlstyle{rm}\Url}\fi

\bibitem[{Anthropic}(2025)]{anthropic2025opus}
{Anthropic}.
\newblock {Claude Opus 4.5 System Card Technical Report}, 2025.
\newblock URL \url{https://assets.anthropic.com/m/64823ba7485345a7/Claude-Opus-4-5-System-Card.pdf}.

\bibitem[Bodenreider(2004)]{bodenreider2004unified}
Bodenreider, O.
\newblock {The Unified Medical Language System (UMLS): Integrating Biomedical Terminology}.
\newblock \emph{Nucleic acids research}, 32:\penalty0 D267--D270, 2004.

\bibitem[Chu et~al.(2025)Chu, Zhai, Yang, Tong, Xie, Schuurmans, Le, Levine, and Ma]{chu2025sft}
Chu, T., Zhai, Y., Yang, J., Tong, S., Xie, S., Schuurmans, D., Le, Q.~V., Levine, S., and Ma, Y.
\newblock {SFT Memorizes, RL Generalizes: A Comparative Study of Foundation Model Post-Training}.
\newblock \emph{arXiv preprint arXiv:2501.17161}, 2025.

\bibitem[Cui et~al.(2025)Cui, Yuan, Wang, Wang, Li, He, Fan, Yu, Xu, Chen, Yuan, Chen, Zhang, Lv, Wang, Yao, Han, Peng, Cheng, Liu, Sun, Zhou, and Ding]{CuiPROCESSREWARDS}
Cui, G., Yuan, L., Wang, Z., Wang, H., Li, W., He, B., Fan, Y., Yu, T., Xu, Q., Chen, W., Yuan, J., Chen, H., Zhang, K., Lv, X., Wang, S., Yao, Y., Han, X., Peng, H., Cheng, Y., Liu, Z., Sun, M., Zhou, B., and Ding, N.
\newblock {Process Reinforcement Through Implicit Rewards}.
\newblock \emph{arXiv preprint arXiv:2502.01456}, 2025.

\bibitem[Damani et~al.(2025)Damani, Puri, Slocum, Shenfeld, Choshen, Kim, and Andreas]{DamaniBEYONDUNCERTAINTY}
Damani, M., Puri, I., Slocum, S., Shenfeld, I., Choshen, L., Kim, Y., and Andreas, J.
\newblock {Beyond Binary Rewards: Training {LM}s to Reason About Their Uncertainty}.
\newblock \emph{arXiv preprint arXiv:2507.16806}, 2025.

\bibitem[Das et~al.(2017)Das, Dhuliawala, Zaheer, Vilnis, Durugkar, Krishnamurthy, Smola, and McCallum]{DasGOLEARNING}
Das, R., Dhuliawala, S., Zaheer, M., Vilnis, L., Durugkar, I., Krishnamurthy, A., Smola, A., and McCallum, A.
\newblock {Go for a Walk and Arrive at the Answer: Reasoning Over Paths in Knowledge Bases Using Reinforcement Learning}.
\newblock \emph{arXiv preprint arXiv:1711.05851}, 2017.

\bibitem[Dedhia et~al.(2025)Dedhia, Kansal, and Jha]{Dedhia2025Bottom-upNeed}
Dedhia, B., Kansal, Y., and Jha, N.~K.
\newblock {Bottom-up Domain-specific Superintelligence: {A} Reliable Knowledge Graph is What We Need}.
\newblock \emph{arXiv preprint arXiv:2507.13966}, 7 2025.

\bibitem[Fodor(1975)]{fodor1975language}
Fodor, J.~A.
\newblock \emph{{The Language of Thought}}, volume~5.
\newblock Harvard University Press, 1975.

\bibitem[{Google DeepMind}(2025)]{GoogleDeepMindGemini3Pro}
{Google DeepMind}.
\newblock {Gemini 3 Pro Model Card}, November 2025.
\newblock URL \url{https://storage.googleapis.com/deepmind-media/Model-Cards/Gemini-3-Pro-Model-Card.pdf}.
\newblock Technical Report.

\bibitem[Gu et~al.(2025)Gu, Fu, Liu, Valanarasu, Codella, Tan, Liu, Jin, Zhang, Wang, et~al.]{gu2025illusion}
Gu, Y., Fu, J., Liu, X., Valanarasu, J. M.~J., Codella, N.~C., Tan, R., Liu, Q., Jin, Y., Zhang, S., Wang, J., et~al.
\newblock {The Illusion of Readiness in Health {AI}}.
\newblock \emph{arXiv preprint arXiv:2509.18234}, 2025.

\bibitem[Gunjal et~al.(2025)Gunjal, Wang, Lau, Nath, Liu, and Hendryx]{GunjalRubrics}
Gunjal, A., Wang, A., Lau, E., Nath, V., Liu, B., and Hendryx, S.
\newblock {Rubrics as Rewards: Reinforcement Learning Beyond Verifiable Domains}.
\newblock \emph{arXiv preprint arXiv:2507.17746}, 2025.

\bibitem[Guo et~al.(2025)Guo, Yang, Zhang, Song, Zhang, Xu, Zhu, Ma, Wang, Bi, et~al.]{guo2025deepseek}
Guo, D., Yang, D., Zhang, H., Song, J., Zhang, R., Xu, R., Zhu, Q., Ma, S., Wang, P., Bi, X., et~al.
\newblock {Deepseek-R1: Incentivizing Reasoning Capability in LLMs Via Reinforcement Learning}.
\newblock \emph{arXiv preprint arXiv:2501.12948}, 2025.

\bibitem[Hu et~al.(2022)Hu, Shen, Wallis, Allen-Zhu, Li, Wang, Wang, Chen, et~al.]{hu2022lora}
Hu, E.~J., Shen, Y., Wallis, P., Allen-Zhu, Z., Li, Y., Wang, S., Wang, L., Chen, W., et~al.
\newblock {LoRA: Low-Rank Adaptation of Large Language Models}.
\newblock \emph{In Proceedings of International Conference on Learning Representations}, 1\penalty0 (2):\penalty0 3, 2022.

\bibitem[Jin et~al.(2025)Jin, Luan, Lyu, Rabusseau, Rabbany, Precup, and Hamdaqa]{jin2025rl}
Jin, H., Luan, S., Lyu, S., Rabusseau, G., Rabbany, R., Precup, D., and Hamdaqa, M.
\newblock {RL Fine-Tuning Heals OOD Forgetting In SFT}.
\newblock \emph{arXiv preprint arXiv:2509.12235}, 2025.

\bibitem[Kamp \& Partee(1995)Kamp and Partee]{kamp1995prototype}
Kamp, H. and Partee, B.
\newblock {Prototype Theory and Compositionality}.
\newblock \emph{Cognition}, 57\penalty0 (2):\penalty0 129--191, 1995.

\bibitem[Kang et~al.(2025)Kang, Kuchnik, Padthe, Vlastelica, Jia, Wu, and Ardalani]{kang2025quagmires}
Kang, F., Kuchnik, M., Padthe, K., Vlastelica, M., Jia, R., Wu, C.-J., and Ardalani, N.
\newblock {Quagmires in SFT-RL Post-Training: When High SFT Scores Mislead and What to Use Instead}.
\newblock \emph{arXiv preprint arXiv:2510.01624}, 2025.

\bibitem[Khatwani et~al.(2025)Khatwani, Cheng, Afshar, Dligach, and Gao]{khatwani2025brittleness}
Khatwani, S., Cheng, H., Afshar, M., Dligach, D., and Gao, Y.
\newblock {Brittleness and Promise: Knowledge Graph Based Reward Modeling for Diagnostic Reasoning}.
\newblock \emph{arXiv preprint arXiv:2509.18316}, 2025.

\bibitem[Kim et~al.(2025)Kim, Jeong, Chen, Li, Lu, Alhamoud, Mun, Grau, Jung, Gameiro, Fan, Park, Lin, Yoon, Yoon, Sap, Tsvetkov, Liang, Xu, Liu, McDuff, Lee, Park, Tulebaev, and Breazeal]{KimMedicalHealthcare}
Kim, Y., Jeong, H., Chen, S., Li, S.~S., Lu, M., Alhamoud, K., Mun, J., Grau, C., Jung, M., Gameiro, R., Fan, L., Park, E., Lin, T., Yoon, J., Yoon, W., Sap, M., Tsvetkov, Y., Liang, P., Xu, X., Liu, X., McDuff, D., Lee, H., Park, H.~W., Tulebaev, S., and Breazeal, C.
\newblock {Medical Hallucination in Foundation Models and Their Impact on Healthcare}.
\newblock \emph{arXiv preprint arXiv:2503.05777}, 2025.

\bibitem[Lightman et~al.(2023)Lightman, Kosaraju, Burda, Edwards, Baker, Lee, Leike, Schulman, Sutskever, and Cobbe]{LightmanLetsStep}
Lightman, H., Kosaraju, V., Burda, Y., Edwards, H., Baker, B., Lee, T., Leike, J., Schulman, J., Sutskever, I., and Cobbe, K.
\newblock {Let's Verify Step by Step}.
\newblock \emph{arXiv preprint arXiv:2305.20050}, 2023.

\bibitem[Lin et~al.(2018)Lin, Socher, and Xiong]{lin2018multi}
Lin, X.~V., Socher, R., and Xiong, C.
\newblock {Multi-Hop Knowledge Graph Reasoning With Reward Shaping}.
\newblock \emph{arXiv preprint arXiv:1808.10568}, 2018.

\bibitem[Matsutani et~al.(2025)Matsutani, Takashiro, Minegishi, Kojima, Iwasawa, and Matsuo]{matsutani2025rl}
Matsutani, K., Takashiro, S., Minegishi, G., Kojima, T., Iwasawa, Y., and Matsuo, Y.
\newblock {RL Squeezes, SFT Expands: A Comparative Study of Reasoning LLMs}.
\newblock \emph{arXiv preprint arXiv:2509.21128}, 2025.

\bibitem[Muennighoff et~al.(2025)Muennighoff, Yang, Shi, Li, Fei-Fei, Hajishirzi, Zettlemoyer, Liang, Cand{\`e}s, and Hashimoto]{muennighoff2025s1}
Muennighoff, N., Yang, Z., Shi, W., Li, X.~L., Fei-Fei, L., Hajishirzi, H., Zettlemoyer, L., Liang, P., Cand{\`e}s, E., and Hashimoto, T.~B.
\newblock {s1: Simple Test-Time Scaling}.
\newblock In \emph{Proceedings of the Conference on Empirical Methods in Natural Language Processing}, pp.\  20286--20332, 2025.

\bibitem[{OpenAI}(2025)]{OpenAIGPT52}
{OpenAI}.
\newblock {Introducing {GPT}-5.2}.
\newblock \url{https://openai.com/index/introducing-gpt-5-2/}, 2025.
\newblock Accessed: 2026.

\bibitem[Organization(1992)]{whoICD10}
Organization, W.~H.
\newblock \emph{{International Statistical Classification of Diseases and Related Health Problems 10th Revision (ICD-10)}}.
\newblock World Health Organization, 1992.

\bibitem[Ouyang et~al.(2022)Ouyang, Wu, Jiang, Almeida, Wainwright, Mishkin, Zhang, Agarwal, Slama, Ray, et~al.]{ouyang2022training}
Ouyang, L., Wu, J., Jiang, X., Almeida, D., Wainwright, C., Mishkin, P., Zhang, C., Agarwal, S., Slama, K., Ray, A., et~al.
\newblock {Training Language Models to Follow Instructions with Human Feedback}.
\newblock \emph{Advances in Neural Information Processing Systems}, 35:\penalty0 27730--27744, 2022.

\bibitem[Rafailov et~al.(2023)Rafailov, Sharma, Mitchell, Ermon, Manning, and Finn]{Rafailov2023DirectModel}
Rafailov, R., Sharma, A., Mitchell, E., Ermon, S., Manning, C.~D., and Finn, C.
\newblock {Direct Preference Optimization: Your Language Model is Secretly a Reward Model}.
\newblock \emph{Advances in Neural Information Processing Systems}, 36, 2023.

\bibitem[Rajani et~al.(2025)Rajani, Gema, Goldfarb-Tarrant, and Titov]{Rajani2025}
Rajani, N., Gema, A.~P., Goldfarb-Tarrant, S., and Titov, I.
\newblock {Scalpel vs. Hammer: {GRPO} Amplifies Existing Capabilities, {SFT} Replaces Them}.
\newblock \emph{arXiv preprint}, 2025.
\newblock Preprint.

\bibitem[Schulman et~al.(2017)Schulman, Wolski, Dhariwal, Radford, and Klimov]{schulman2017proximal}
Schulman, J., Wolski, F., Dhariwal, P., Radford, A., and Klimov, O.
\newblock {Proximal Policy Optimization Algorithms}.
\newblock \emph{arXiv preprint arXiv:1707.06347}, 2017.

\bibitem[Shrivastava et~al.(2025)Shrivastava, Awadallah, Balachandran, Garg, Behl, and Papailiopoulos]{Shrivastava2025SampleReasoning}
Shrivastava, V., Awadallah, A., Balachandran, V., Garg, S., Behl, H., and Papailiopoulos, D.
\newblock {Sample More to Think Less: Group Filtered Policy Optimization for Concise Reasoning}.
\newblock \emph{arXiv preprint arXiv:2508.09726}, 2025.

\bibitem[Wang et~al.(2025)Wang, Li, Sun, Chen, Liu, Wu, Lu, Song, and Abbasi-Yadkori]{Wang2025HierarchicalModel}
Wang, G., Li, J., Sun, Y., Chen, X., Liu, C., Wu, Y., Lu, M., Song, S., and Abbasi-Yadkori, Y.
\newblock {Hierarchical Reasoning Model}.
\newblock \emph{arXiv preprint arXiv:2506.21734}, 2025.

\bibitem[Wang et~al.(2024)Wang, Chen, Hu, Yang, Liu, Shen, Wei, Zhang, Gu, Zhou, Pan, Zhang, and Chen]{Wang2024}
Wang, J., Chen, M., Hu, B., Yang, D., Liu, Z., Shen, Y., Wei, P., Zhang, Z., Gu, J., Zhou, J., Pan, J.~Z., Zhang, W., and Chen, H.
\newblock {Learning to Plan for Retrieval-Augmented Large Language Models from Knowledge Graphs}.
\newblock In \emph{Findings of the Association for Computational Linguistics: EMNLP 2024}, pp.\  7813--7835. Association for Computational Linguistics, 2024.

\bibitem[Wei et~al.(2022)Wei, Wang, Schuurmans, Bosma, Xia, Chi, Le, Zhou, et~al.]{wei2022chain}
Wei, J., Wang, X., Schuurmans, D., Bosma, M., Xia, F., Chi, E., Le, Q.~V., Zhou, D., et~al.
\newblock {Chain-of-Thought Prompting Elicits Reasoning in Large Language Models}.
\newblock \emph{Advances in Neural Information Processing Systems}, 35:\penalty0 24824--24837, 2022.

\bibitem[Weng(2024)]{RewardLilLog}
Weng, L.
\newblock {Reward Hacking in Reinforcement Learning}.
\newblock \url{https://lilianweng.github.io/posts/2024-11-28-reward-hacking/}, 2024.
\newblock Lil'Log blog post.

\bibitem[Xie et~al.(2025)Xie, Gao, Ren, Luo, Hong, Dai, Zhou, Qiu, Wu, and Luo]{XieLogicRL}
Xie, T., Gao, Z., Ren, Q., Luo, H., Hong, Y., Dai, B., Zhou, J., Qiu, K., Wu, Z., and Luo, C.
\newblock {Logic-{RL}: Unleashing {LLM} Reasoning with Rule-Based Reinforcement Learning}.
\newblock \emph{arXiv preprint arXiv:2502.14768}, 2025.

\bibitem[Xiong et~al.(2017)Xiong, Hoang, and Wang]{xiong2017deeppath}
Xiong, W., Hoang, T., and Wang, W.~Y.
\newblock {DeepPath: A Reinforcement Learning Method for Knowledge Graph Reasoning}.
\newblock \emph{arXiv preprint arXiv:1707.06690}, 2017.

\bibitem[Yan et~al.(2025)Yan, Tang, Guan, Wang, Wang, Liu, Yang, and Jiang]{YanRLKGF}
Yan, L., Tang, C., Guan, Y., Wang, H., Wang, S., Liu, H., Yang, Y., and Jiang, J.
\newblock {{RLKGF}: Reinforcement Learning from Knowledge Graph Feedback without Human Annotations}.
\newblock In \emph{Findings of the Association for Computational Linguistics: ACL 2025}, pp.\  6619--6633, 2025.

\bibitem[Yang et~al.(2025)Yang, Li, Yang, Zhang, Hui, Zheng, Yu, Gao, Huang, Lv, et~al.]{yang2025qwen3}
Yang, A., Li, A., Yang, B., Zhang, B., Hui, B., Zheng, B., Yu, B., Gao, C., Huang, C., Lv, C., et~al.
\newblock {Qwen3 Technical Report}.
\newblock \emph{arXiv preprint arXiv:2505.09388}, 2025.

\bibitem[Yasunaga et~al.(2021)Yasunaga, Ren, Bosselut, Liang, and Leskovec]{yasunaga2021qa}
Yasunaga, M., Ren, H., Bosselut, A., Liang, P., and Leskovec, J.
\newblock {QA-GNN: Reasoning with Language Models and Knowledge Graphs for Question Answering}.
\newblock \emph{arXiv preprint arXiv:2104.06378}, 2021.

\bibitem[Yin et~al.(2025)Yin, Qu, Yang, Cong, and Wang]{Yin2025TowardLearning}
Yin, M., Qu, Y., Yang, L., Cong, L., and Wang, M.
\newblock {Toward Scientific Reasoning in {LLM}s: Training from Expert Discussions via Reinforcement Learning}.
\newblock \emph{arXiv preprint arXiv:2505.19501}, 2025.

\bibitem[Yuan et~al.(2025)Yuan, Chen, Zhang, Cui, Wang, You, Ding, Liu, Sun, and Peng]{yuan2025f}
Yuan, L., Chen, W., Zhang, Y., Cui, G., Wang, H., You, Z., Ding, N., Liu, Z., Sun, M., and Peng, H.
\newblock {From $ f (x) $ and $ g (x) $ to $ f (g (x)) $: LLMs Learn New Skills in RL by Composing Old Ones}.
\newblock \emph{arXiv preprint arXiv:2509.25123}, 2025.

\bibitem[Yue et~al.(2025)Yue, Chen, Lu, Zhao, Wang, Song, and Huang]{yue2025does}
Yue, Y., Chen, Z., Lu, R., Zhao, A., Wang, Z., Song, S., and Huang, G.
\newblock {Does Reinforcement Learning Really Incentivize Reasoning Capacity in LLMs Beyond the Base Model?}
\newblock \emph{arXiv preprint arXiv:2504.13837}, 2025.

\bibitem[Zhang et~al.(2025)Zhang, Zheng, Wu, Zhang, Lin, Yu, Liu, Zhou, and Lin]{Zhang2025TheReasoning}
Zhang, Z., Zheng, C., Wu, Y., Zhang, B., Lin, R., Yu, B., Liu, D., Zhou, J., and Lin, J.
\newblock {The Lessons of Developing Process Reward Models in Mathematical Reasoning}.
\newblock \emph{arXiv preprint arXiv:2501.07301}, 2025.

\bibitem[Zhu et~al.(2025)Zhu, Xia, Wei, Chen, Chen, and Meng]{Zhu2025}
Zhu, X., Xia, M., Wei, Z., Chen, W.-L., Chen, D., and Meng, Y.
\newblock {The Surprising Effectiveness of Negative Reinforcement in {LLM} Reasoning}.
\newblock \emph{arXiv preprint arXiv:2506.01347}, 2025.

\end{thebibliography}
\bibliographystyle{icml2026}

%%%%%%%%%%%%%%%%%%%%%%%%%%%%%%%%%%%%%%%%%%%%%%%%%%%%%%%%%%%%%%%%%%%%%%%%%%%%%%%
%%%%%%%%%%%%%%%%%%%%%%%%%%%%%%%%%%%%%%%%%%%%%%%%%%%%%%%%%%%%%%%%%%%%%%%%%%%%%%%
% APPENDIX
%%%%%%%%%%%%%%%%%%%%%%%%%%%%%%%%%%%%%%%%%%%%%%%%%%%%%%%%%%%%%%%%%%%%%%%%%%%%%%%
%%%%%%%%%%%%%%%%%%%%%%%%%%%%%%%%%%%%%%%%%%%%%%%%%%%%%%%%%%%%%%%%%%%%%%%%%%%%%%%
\newpage
\appendix
\onecolumn

\section{Zero-RL and Reward Ablation Studies on Qwen3 8B}
\label{Appendix A}

In this section, we provide the empirical details of our experiments with the ``Zero-RL" approach (applying RL directly to the base model without a prior SFT phase). These ablations were instrumental in determining our final training scale, pipeline design, and reward configuration. 

\subsection{Zero-RL Performance and Scaling}
We evaluate the performance of the Qwen3 8B base model using the GRPO algorithm across three data scales: 5k, 10k, and 24.66k examples. For each scale, we test four reward configurations: (1) path alignment (path overlap), (2) Jaccard similarity (distillation), (3) binary correctness, and (4) all rewards combined (see Table \ref{tab:reward_ablation}). All experiments are performed on the Qwen3 8B model using 8 H100 NVIDIA GPUs.

Note that we always include the binary correctness reward as a minimal signal.

\begin{table}[ht]
\caption{Ablation study on reward components across training scales on Qwen3 8B.}
\label{tab:reward_ablation}
\vskip 0.15in
\begin{center}
\begin{small}
\begin{sc}
\begin{tabular}{rccc}
\toprule
Training Size & Reward & Accuracy & $\sim$Training Time (hrs) \\
\midrule
0 & Baseline & 64.98\% & -- \\
\midrule
\multirow{3}{*}{5k} 
  & Path Align. & 67.51\% & 12 \\
  & Jaccard Sim. & 68.03\% & 12 \\
  & Binary & \textbf{69.36\%} & 12 \\
\midrule
\multirow{4}{*}{10k} 
  & Path Align. & 64.24\% & 24 \\
  & Jaccard Sim. & 65.74\% & 24 \\
  & Binary & 67.56\% & 24 \\
  & All Rewards & \textbf{68.44\%} & 24 \\
\midrule
\multirow{4}{*}{24.66k} 
  & Path Align. & 68.41\% & 65 \\
  & Jaccard Sim. & 65.77\% & 65 \\
  & Binary & \textbf{70.18\%} & 65 \\
  & All Rewards & 68.44\% & 65 \\
\bottomrule
\end{tabular}
\end{sc}
\end{small}
\end{center}
\vskip -0.1in
\end{table}

Key Observations:
\begin{itemize}
    \item \textbf{SFT Requirement:} Whereas Zero-RL provides a marginal improvement over the base model, no configuration reaches the performance levels of the SFT-only baseline (70.86\%) or our final SFT+RL model ($\sim$ 82\%). This confirms that the model requires a grounded factual foundation from SFT before it can effectively leverage RL for compositional reasoning. In the final SFT+RL pipeline, path alignment, along with binary correctness (with upweighted negative reward), provides the best and most reliable signal.

    \item \textbf{Scale vs. Efficiency:} Performance gains from 5k to 24.66k examples are minimal (e.g., a $\sim$0.8\% gain for binary correctness reward) while significantly increasing training time from 12 to 65 hours. This justifies our decision to set the RL stage to use a high-quality 5k subset of our final pipeline.

    \item \textbf{Reward Stability:} The binary correctness reward remains the most stable signal in the Zero-SFT setting. The distillation-based rewards are prone to reward hacking and stylistic mimicry without providing superior reasoning abilities.
\end{itemize}

\subsection{GRPO on LoRA Parameters Only}
We also explore a variant of the pipeline where we perform SFT followed by GRPO, with updates limited only to the LoRA modules. This configuration yields an accuracy of 66.75\% on the 8B model. The significantly lower performance compared to our full SFT+RL results suggests that compositional reasoning requires more extensive/aggressive weight updates during RL than those afforded by the restricted LoRA-only approach.

\newpage
\section{SFT+RL Ablation Studies on Qwen3 8B}
\label{Appendix B}
This study investigates the synergy between the binary reward and path alignment reward, and provides the final missing link by ablating the specific reward components used in the RL stage. Whereas Appendix \ref{Appendix A} focuses on the Zero-RL settings, the following experiments demonstrate the impact of reward designs on the model that has already undergone SFT. We conduct these ablations on the Qwen3 8B model to efficiently identify the optimal configuration for the larger 14B SFT+RL pipeline.

We specifically evaluate the addition of negative reinforcement \cite{Zhu2025} by penalizing the model more for incorrect final answers than correct reasoning paths. As shown in Table \ref{tab:sft_rl_ablation}, the combination of path-derived signals and negative binary reward provides the most robust reasoning improvements, significantly outperforming configurations that rely on binary signals alone or a combination of all rewards.

\begin{table}[ht]
\caption{Ablation study on reward components with the SFT+RL training pipeline (8B Model). All RL runs are conducted for 5k steps following a 19.66k SFT baseline.}
\label{tab:sft_rl_ablation}
\vskip 0.15in
\begin{center}
\begin{small}
\begin{sc}
\begin{tabular}{llc}
\toprule
Training Setup & Reward Configuration & Accuracy (\%) \\
\midrule
SFT Baseline  & - & 70.86\% \\
\midrule
\multirow{5}{*}{19.66k SFT + 5k RL} 
  & Path Alignment Only & 79.29\% \\
  & Path Alignment + Binary & 68.03\% \\
  & Binary Only (Normal) & 79.46\% \\
  & Negative Binary Only & 79.54\% \\
  & All Rewards & 55.21\% \\
  & \textbf{Path Align. + Negative Binary} & \textbf{82.20\%} \\
\bottomrule
\end{tabular}
\end{sc}
\end{small}
\end{center}
\vskip -0.1in
\end{table}

Key Insights:
\begin{itemize}
    \item \textbf{The Synergy of Negative Reinforcement:} Consistent with the findings of \cite{Zhu2025}, we observe that ``Normal" binary rewards (positive reinforcement only) can be unstable when combined with path-aligned rewards. Transitioning to negative binary rewards while rewarding valid intermediate paths yields our highest accuracy of 82.20\%.

    \item \textbf{The ``All Rewards" Failure:} Attempting to combine all reward signals (path alignment, Jaccard similarity, normal binary, and thinking quality) leads to a significant performance collapse (55.21\%). This suggests that over-optimizing the reward signal can lead to conflicting gradients or reward hacking, where the model fails to optimize for the primary reasoning task.

    \item \textbf{Effect of the Path Alignment Signal:} Even without outcome-based binary rewards, path alignment only provides a substantial $\sim$9\% boost over the SFT baseline. This confirms that the KG itself acts as a sufficient implicit reward model to guide the model towards better compositional reasoning.
    
\end{itemize}

\newpage
\section{Additional Reward Function Formulations}
\label{Appendix C}
In addition to the path-alignment reward, $R_{path}$, and the binary correctness reward, $R_{bin}$, we investigate two alternative reward functions for RL experiments: semantic answer similarity (distillation-based reward) ($R_{sim}$) and thinking quality ($R_{think}$). Whereas these functions provided some learning signals, they are ultimately less effective than path-derived rewards in enabling deep compositional reasoning.

\subsection{Semantic Answer Similarity Reward $(R_{sim})$}
The semantic answer similarity (or distillation) reward, $R_{sim}$, is designed to distill the reasoning patterns of the SFT model by rewarding semantic overlap between the model-generated reasoning and the ground truth answer reasoning trace.

The reward is calculated using the Jaccard similarity of normalized token sets:
\[
R_{sim} = \frac{|T_{model} \cap T_{target}|}{|T_{model} \cup T_{target}|} \times \phi_{rep},
\]
where $T_{model}$ is the set of unique tokens from the thinking trace of the model, $T_{target}$ is the token set of the ground truth reasoning text (distilled from a larger LLM), and $\phi_{rep}$ is a repetition penalty factor used to discourage repetitive generation. This function encourages the model to mention the correct clinical entities and mechanisms associated with the target answer, though it does not explicitly verify the logical validity of the connections between those entities.

\subsection{Thinking Quality Reward $(R_{think})$}
The thinking quality reward, $R_{think}$, is designed to encourage the model to produce well-organized, stepwise reasoning. It evaluates the response based on its adherence to logical formatting cues. It is calculated as a weighted sum of three structural components:
\begin{itemize}
    \item \textbf{Structure (50\%):} A binary indicator of whether the reasoning trace exceeds a minimum length threshold (20 characters).
    \item \textbf{Step-wise Keywords (30\%):} A normalized count of logical-thinking indicator words such as ``first," ``then," ``therefore," and ``finally."
    \item \textbf{Enumeration (20\%):} A score based on the presence of numbered lists (e.g., ``1.," ``2."), which often indicate a systematic breakdown of a multi-hop clinical diagnostic problem.
\end{itemize}

In addition, we employ a penalty, $\phi_{rep}$, to discourage repetitive output. Whereas $R_{think}$ shows initial promise in improving the readability of the model outputs, it often leads to reward hacking in contrast to the rigorous logical traversal induced by path-derived reward signals.

\newpage
\section{Train-Test Split Overlap Analysis}
\label{Appendix D}

A critical concern in any reasoning-based benchmark is the potential for data leakage or memorization. As per the design of our data curation pipeline, which builds upon the methodology established in \cite{Dedhia2025Bottom-upNeed}, each question created by traversing a KG path is unique. Per design, arriving at a correct answer requires explicit reasoning involving all intermediate edges and nodes in the path, rather than simple pattern matching.

\begin{figure}[h!]
  \vskip -0.5em
  \centering
  \begin{minipage}{0.48\columnwidth}
    \centering
    \includegraphics[width=\textwidth]{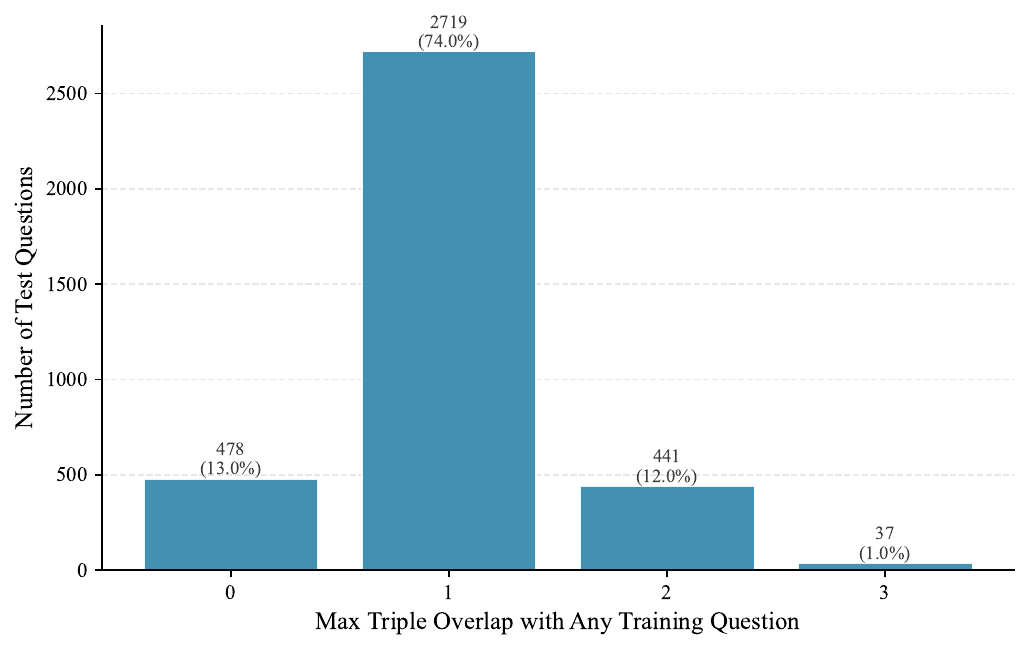}
    \caption{\textbf{Number of KG Triples shared with the Training Set}}
    \label{kg_overlap}
  \end{minipage}
  \hfill
  \begin{minipage}{0.48\columnwidth}
    \centering
    \includegraphics[width=\textwidth]{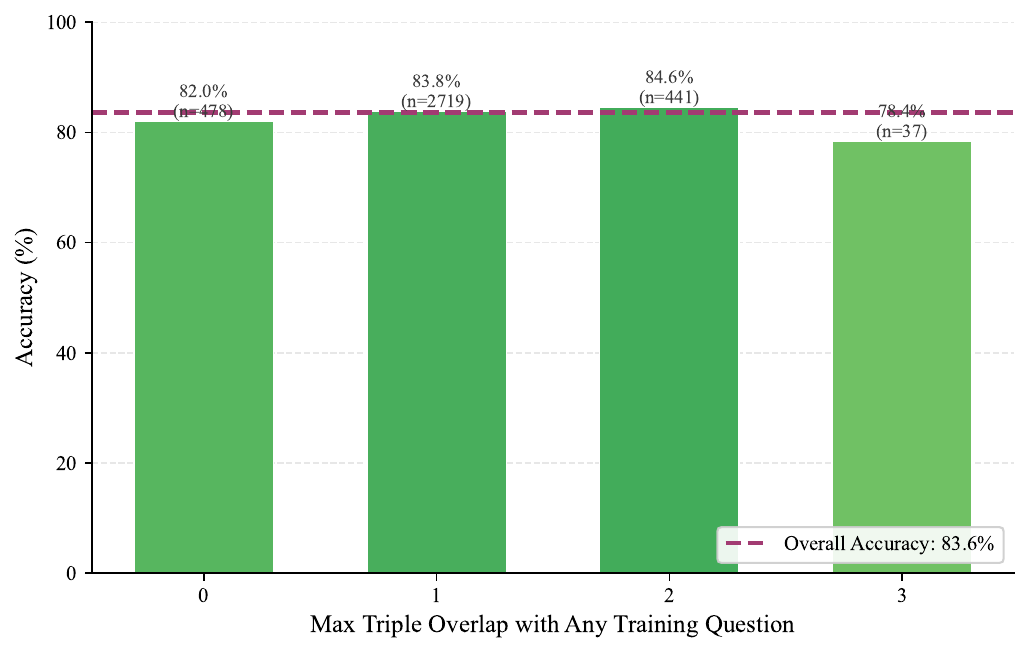}
    \caption{\textbf{Accuracy by Number of Overlapping KG Triples}}
    \label{kg_overlap_acc}
  \end{minipage}
  \vskip -0.5em
\end{figure}

We intentionally allow partial overlap of individual nodes or single triples between the training and test sets to ensure the benchmark exercises broad coverage of medical concepts. To more concretely address concerns regarding the influence of partial knowledge overlap, we performed a rigorous analysis of our model performance on the test set categorized by the longest consecutive matching subsequence of triples found in any training question. This match can start at any position in either the training or test path. The distribution and corresponding accuracy metrics are detailed in Figs.~\ref{kg_overlap}, \ref{kg_overlap_acc}, respectively.

Only 1.0\% of test questions have a full 3-triple chain match. Since our training set is restricted to a maximum of 3-hop reasoning tasks, this ensures that the vast majority of our benchmark consists of structurally unique reasoning tasks. Furthermore, there is no clear trend in accuracy as overlap increases. This suggests that performance is not significantly predicted by the presence of familiar triple sequences. Interestingly, the 3-overlap category exhibits slightly lower accuracy (78.4\%) than the 0-overlap category (82.0\%). This further demonstrates that our model internalizes the logic of composition rather than memorizing specific paths.

Whereas the majority of questions share at least one triple with the training set, note that the ICD-Bench test set contains only 2-5-hop questions. This requires the model to perform novel composition even when individual building blocks have been previously seen in different contexts.

% \begin{figure}[H]
%   \vskip -0.5em  % negative space instead of positive
%   \begin{center}
%     \centerline{\includegraphics[width=0.7\textwidth]{figs/accuracy_by_triple_overlap.pdf}}
%     \caption{
%       \textbf{Accuracy by Number of Overlapping KG Triples}
%     }
%     \label{kg_overlap_acc}
%   \end{center}
%   \vskip -0.5em  % add negative space after too if needed
% \end{figure}

\newpage
\section{Training Hyperparameters and GRPO Configuration}
\label{Appendix E}
In this section, we detail the specific hyperparameter configurations used for both the SFT (Table \ref{tab:sft_hyperparams}) and RL (GRPO) (Table \ref{tab:grpo_hyperparams}) stages of our pipeline.

\subsection{Hardware Compute}
All experiments are conducted using high-performance GPU clusters. The 8B model runs are executed on a node equipped with 8$\times$ NVIDIA H100 GPUs. For the 14B parameter model, we use 8$\times$ NVIDIA H200 GPUs to accommodate for increased memory requirements and ensure efficient throughput in the RL phase. DeepSpeed is employed as the inference engine to optimize model sharding, memory management, and logging during GRPO.

\subsection{SFT Stage}
For the SFT stage, we use the LoRA method \cite{hu2022lora}. The hyperparameters are kept constant across both model scales. 

\begin{table}[ht]
\caption{Hyperparameters for SFT.}
\label{tab:sft_hyperparams}
\vskip 0.15in
\begin{center}
\begin{small}
\begin{sc}
\begin{tabular}{lc}
\toprule
Hyperparameter & Value \\
\midrule
LoRA Rank ($r$) & 16 \\
LoRA Alpha ($\alpha$) & 16 \\
LoRA Dropout & 0.05 \\
Learning Rate & $2 \times 10^{-4}$ \\
\bottomrule
\end{tabular}
\end{sc}
\end{small}
\end{center}
\vskip -0.1in
\end{table}

\subsection{RL (GRPO) Stage}
Given the high cost of model generation in full-parameter GRPO, we optimize per-device efficiency while maintaining a stable learning signal by using a constant learning rate with a warmup period. The choice of $\text{num\_generations} = 2$ is made to balance memory requirements with completion length. We found that low temperature is essential for maintaining logical consistency and diversity.

\begin{table}[H]
\caption{Hyperparameters for GRPO training.}
\label{tab:grpo_hyperparams}
\vskip 0.15in
\begin{center}
\begin{small}
\begin{sc}
\begin{tabular}{lc}
\toprule
Hyperparameter & Value \\
\midrule
Num Generations ($G$) & 2 \\
Batch Size (per device) & 1 \\
Learning Rate & $8 \times 10^{-6}$ \\
LR Schedule & Constant w/ Warmup \\
Temperature ($T$) & 0.6 \\
Top-p & 0.9 \\
Optimizer & AdamW \\
Precision & BF16 \\
Max Completion Length & 1792 \\
Repetition Penalty & 1.15 \\
\bottomrule
\end{tabular}
\end{sc}
\end{small}
\end{center}
\vskip -0.1in
\end{table}

\newpage
\section{Sample Model Responses}
\label{Appendix F}
Next, we present representative samples of the output of our 14B SFT+RL model on complex, multi-hop medical queries from the ICD-Bench test set. These examples illustrate how our model effectively composes several 1-hop facts to solve a 3-hop and a 5-hop problem, respectively. \\

\definecolor{medicalblue}{RGB}{30, 80, 150}
\definecolor{thoughtback}{RGB}{245, 250, 255}
\definecolor{pathhighlight}{RGB}{255, 230, 180}
\definecolor{truepathback}{RGB}{210, 245, 210}

\begin{tcolorbox}[
    colframe=medicalblue,
    colback=white,
    title={\textbf{Detailed Compositional Reasoning | Category: Neoplasms | Difficulty: 1 | Num Hops: 3}},
    colbacktitle=medicalblue,
    fonttitle=\sffamily\bfseries,
    sharp corners,
    boxrule=1pt,
    enhanced,
    breakable,
    attach boxed title to top left={yshift=-2mm, xshift=2mm},
    boxed title style={sharp corners, colframe=medicalblue}
]
    \small
    \textbf{Question:} 
    A 6-month-old infant presents with a rapidly growing mass in their left flank. Imaging reveals a large renal mass suspicious for atypical teratoid rhabdoid tumor (ATRT). Family history is significant for a maternal grandfather with multiple cutaneous neurilemmomas diagnosed in his 60s. Genetic testing confirms a germline pathogenic variant in \textbf{SMARCB1} in the infant. Which of the following secondary findings would warrant the MOST urgent investigation in this patient?

    \vspace{0.2cm}
    \textbf{Options:}
    \begin{itemize}[leftmargin=*, nosep]
        \item[A.] Mild macrocephaly with normal development.
        \item[B.] Small, subcutaneous lipomas noted on physical exam.
        \item[C.] Bilateral sensorineural hearing loss identified on newborn screening.
        \item[D.] Widening of the mediastinum on chest X-ray following a persistent cough.
    \end{itemize}

    \vspace{0.4cm}
    \begin{tcolorbox}[
        colback=thoughtback,
        colframe=medicalblue!30,
        title=Step-by-Step Reasoning ($<think>$),
        fonttitle=\small\sffamily\bfseries,
        coltitle=medicalblue,
        boxrule=0.5pt,
        left=2mm, right=2mm, top=1mm, bottom=1mm
    ]
        \footnotesize
        Okay, let's break this down. We have a 6-month-old baby with a likely ATRT, a pretty aggressive brain tumor in this context, and the genetic testing confirms a germline \textit{SMARCB1} mutation. This mutation is associated with \colorbox{pathhighlight}{Rhabdoid Tumor Predisposition Syndrome (RTPS),} which significantly increases the risk of certain cancers, including the ATRT seen here.

        Now we need to look at the secondary findings and figure out which one needs the most urgent attention. Let's go through the options:

        \begin{itemize}[leftmargin=*, nosep]
            \item \textbf{A. Mild macrocephaly with normal development:} While macrocephaly can be seen in some genetic syndromes, including RTPS, mild macrocephaly isn't typically an immediate life-threatening sign, especially when development is normal. It's something to note but might not be the \textbf{most} urgent thing to investigate right now compared to other possibilities.
            \item \textbf{B. Small, subcutaneous lipomas:} Like macrocephaly, lipomas can occur in RTPS, but small lipomas are generally benign and don't usually require urgent evaluation unless they are growing rapidly or causing symptoms. They don't point to an immediate, critical problem.
            \item \textbf{C. Bilateral sensorineural hearing loss identified on newborn screening:} Hearing loss is a known feature of some SMARCB1 related conditions. However, it's usually detected early (like via newborn screening) and while important for the child's development, it's not typically an immediate medical emergency requiring the same level of urgency as a potential airway or cardiac issue.
            \item \textbf{D. Widening of the mediastinum on chest X-ray following a persistent cough:} This is the big one. A \colorbox{pathhighlight}{widening mediastinum,} especially in the context of a persistent cough, is highly suspicious for a \colorbox{pathhighlight}{thymoma.} Thymomas are tumors located in the anterior mediastinum. While not always malignant, they can be associated with myasthenia gravis, hypogammaglobulinemia, and sometimes other paraneoplastic syndromes. Crucially, large or malignant thymomas can compress airways or the superior vena cava, leading to respiratory distress or superior vena cava syndrome, which is a medical emergency requiring urgent evaluation and potential intervention.
        \end{itemize}

        Considering the infant already has a severe malignancy (ATRT) and a confirmed SMARCB1 mutation, the possibility of a second, potentially rapidly progressive and life-threatening condition like a thymoma, which can cause immediate symptoms (breathing difficulty, facial/arm swelling, neck vein distension), makes the \colorbox{pathhighlight}{mediastinal widening} the most critical finding demanding urgent investigation. The respiratory compromise from a thymoma is a time-sensitive issue.

        Therefore, the \colorbox{pathhighlight}{widening mediastinum} is the most pressing concern that needs investigation urgently.
    \end{tcolorbox}

    \vspace{0.2cm}
\begin{tcolorbox}[
    colback=truepathback,
    colframe=green!50!black,
    boxrule=0.5pt,
    left=2mm, right=2mm, top=1.5mm, bottom=1.5mm
]
    \footnotesize 
    \textbf{Ground Truth Knowledge Graph Path:}
    
    \vspace{0.2cm}
    \centering
    \scriptsize
    \texttt{\colorbox{pathhighlight}{RTPS Type 1}} 
    $\xrightarrow{\text{\tiny may be allelic with}}$ 
    \texttt{\colorbox{pathhighlight}{Schwannomatosis Type 1}}
    $\xrightarrow{\text{\tiny is a risk factor of}}$ 
    \texttt{\colorbox{pathhighlight}{Neurilemmoma}} 
    $\xrightarrow{\text{\tiny may cause}}$ 
    \texttt{\colorbox{pathhighlight}{Mediastinal Enlargement}}
\end{tcolorbox}

    \vspace{0.1cm}
    \textbf{Final Answer: \color{medicalblue}D}\quad \textcolor{green!70!black}{$\checkmark$ Correct}
\end{tcolorbox}

% ANOTHER RESPONSE BOX

\definecolor{medicalblue}{RGB}{30, 80, 150}
\definecolor{thoughtback}{RGB}{245, 250, 255}
\definecolor{pathhighlight}{RGB}{255, 230, 180}
\definecolor{truepathback}{RGB}{210, 245, 210}

\begin{tcolorbox}[
    colframe=medicalblue,
    colback=white,
    title={\textbf{Detailed Compositional Reasoning | Category: Neoplasms | Difficulty: 5 | Num Hops: 5}},
    colbacktitle=medicalblue,
    coltitle=white,
    fonttitle=\sffamily\bfseries,
    sharp corners,
    boxrule=1pt,
    enhanced,
    breakable,
    attach boxed title to top left={yshift=-2mm,xshift=2mm},
    boxed title style={sharp corners,colframe=medicalblue}
]
    \small
    \textbf{Question:} 
    A 17-year-old female presents with worsening pain and swelling in her left distal femur. Imaging reveals a lytic lesion with periosteal reaction highly suggestive of Ewing's sarcoma. She is started on a multi-agent chemotherapy regimen including Etoposide. Several weeks into treatment, she develops significant mucositis and general malaise, leading to difficulty maintaining adequate oral nutrition. Her oncologist is considering adjunctive therapies to improve her nutritional status and muscle mass while carefully considering potential risks and interactions. Which of the following factors should be given MOST consideration when making this decision?

    \vspace{0.2cm}
    \textbf{Options:}
    \begin{itemize}[leftmargin=*, nosep]
        \item[A.] The patient's reported history of childhood asthma and potential for steroid-induced bronchospasm.
        \item[B.] The documented presence of a BRCA1 mutation in a distant relative and the risk of increased malignancy with certain growth factors.
        \item[C.] The potential for teratogenic effects of some anabolic steroids, given the need for effective contraception.
        \item[D.] The need to avoid medications that could exacerbate chemotherapy-induced neuropathy, potentially hindering physical rehabilitation.
    \end{itemize}

    \vspace{0.4cm}
    
    \noindent\colorbox{thoughtback}{%
    \begin{minipage}{\dimexpr\linewidth-4mm}
        \vspace{1mm}
        \footnotesize
        \textbf{Step-by-Step Reasoning ($<think>$):}
        
        \vspace{0.2cm}
        Okay, let's break this down. We have a 17-year-old patient with \colorbox{pathhighlight}{Ewing's sarcoma,} getting chemotherapy that includes \colorbox{pathhighlight}{Etoposide.} She's having side effects like mucositis and general malaise, making it hard to eat. The team is thinking about adding something to help her nutrition and build muscle mass, likely an anabolic steroid or similar agent. We need to figure out which factor is \textit{most} important to consider when choosing this adjunctive therapy.

        \vspace{0.3cm}
        \textbf{1. Understand the Core Problem:} The patient needs help with nutrition and muscle mass due to chemo side effects. Potential solutions might involve anabolic steroids or similar compounds. We need to pick the \textit{most critical} factor influencing this decision.

        \vspace{0.2cm}
        \textbf{2. Analyze Option A (Childhood Asthma):} The patient has childhood asthma. Anabolic steroids, like Oxymetholone, can be used to treat certain types of pulmonary issues, including sometimes asthma. So, there's a potential therapeutic benefit here. However, the primary concern with asthma and steroids is usually bronchospasm, which is a risk with \textit{corticosteroids} (like Prednisone), not typically the main issue directly with \textit{anabolic} steroids, although some anabolic steroids can have corticosteroid effects or cause fluid retention. While important for overall care, it's probably not the \textit{most} critical factor specifically for choosing an adjunctive therapy for nutrition/muscle mass itself, especially compared to direct risks of the therapy.

        \vspace{0.2cm}
        \textbf{3. Analyze Option B (BRCA1 Mutation):} The BRCA1 mutation is relevant for cancer predisposition (increases risk of breast, ovarian cancer). While some growth factors (like Peginesatide) have been associated with risks in BRCA1 mutation carriers, this is a very specific scenario and doesn't directly relate to the common adjunctive agents used for nutritional support or muscle mass building, which are often anabolic steroids. It's highly unlikely to be the \textit{most} critical factor for this patient's specific situation.

        \vspace{0.2cm}
        \textbf{4. Analyze Option D (Neuropathy):} The patient is on Etoposide, which \textit{can} cause neurotoxicity (like peripheral neuropathy). Adding another drug that could potentially worsen this (maybe some agents used for nutrition/muscle mass have neurotoxic potential, or the patient's overall condition makes them more sensitive) is definitely a consideration for quality of life and rehabilitation. It's a valid concern, but is it the \textit{most} critical? Probably not, especially when compared to other potential issues.

        \vspace{0.2cm}
        \textbf{5. Analyze Option C (Teratogenic Effects):} The patient is 17 years old. While not yet of childbearing age, it's crucial to consider future fertility. Anabolic steroids, such as \colorbox{pathhighlight}{Oxandrolone,} are known to be \colorbox{pathhighlight}{teratogenic,} meaning they can cause birth defects. This is a serious risk if the patient were to become \colorbox{pathhighlight}{pregnant,} either during treatment or in the near future after stopping treatment. Given her age and the potential long-term implications of treatment side effects, the risk of teratogenesis stands out as a major, long-term safety concern that \textit{must} be addressed. If an anabolic steroid is considered, effective contraception is absolutely essential, and the teratogenic potential is a primary reason why it needs careful consideration and management. This risk is significant and directly related to the choice of therapy, making it arguably the \textit{most} critical factor to consider.

        \vspace{0.2cm}
        \textbf{6. Synthesize and Conclude:} Comparing the options, the \colorbox{pathhighlight}{teratogenic potential} (Option C) is the most critical factor. Why? Because the patient is young (17), and the potential for future pregnancy exists. Anabolic steroids, which might be considered for her muscle mass and nutrition, are teratogenic. This is a serious long-term health risk that needs careful planning and discussion. While neuropathy (D) is a concern, and asthma (A) has some relevance, the potential for teratogenesis is a major safety issue directly impacting the choice and management of therapy for a young female patient.

        \vspace{0.3cm}
        Therefore, the factor that should be given the MOST consideration is the potential for teratogenic effects.
        \vspace{1mm}
    \end{minipage}%
    }

    \vspace{0.2cm}
    \begin{tcolorbox}[
        colback=truepathback,
        colframe=green!50!black,
        boxrule=0.5pt,
        left=2mm, right=2mm, top=1.5mm, bottom=1.5mm
    ]
        \footnotesize 
        \textbf{Ground Truth Knowledge Graph Path (5 Hops):}
        
        \vspace{0.15cm}
        \centering
        \scriptsize
        \colorbox{pathhighlight}{\texttt{Ewing's sarcoma}} 
        $\xrightarrow{\text{\tiny may treat}}$ 
        \colorbox{pathhighlight}{\texttt{Etoposide}}
        
        \vspace{0.1cm}
        $\xrightarrow{\text{\tiny may treat}}$ 
        \colorbox{pathhighlight}{\texttt{Brain neoplasms}} 
        $\xrightarrow{\text{\tiny may treat}}$ 
        \colorbox{pathhighlight}{\texttt{Hydroxycarbamide}}
        
        \vspace{0.1cm}
        $\xrightarrow{\text{\tiny may contraindicate}}$ 
        \colorbox{pathhighlight}{\texttt{Pregnancy}} 
        $\xrightarrow{\text{\tiny may contraindicate}}$ 
        \colorbox{pathhighlight}{\texttt{Oxandrolone}}
    \end{tcolorbox}

    \vspace{0.1cm}
    \textbf{Final Answer: \color{medicalblue}C} \quad \textcolor{green!70!black}{$\checkmark$ Correct}
\end{tcolorbox}

\newpage
\section{SFT+RL Pipeline Algorithm}
\label{Appendix G}
To provide a clear, formal overview of our training process, we present the SFT+RL Pipeline in Algorithm \ref{alg:pipeline}. This algorithm details the transition from high-quality grounded SFT to path-aligned RL using the GRPO framework.

\begin{algorithm}[h!]
   \caption{Grounded Compositional Reasoning Pipeline (SFT+RL)}
   \label{alg:pipeline}
\begin{algorithmic}
   \STATE {\bfseries Input:} Base model $\theta_{base}$, Knowledge graph $\mathcal{G}$, Training set $\mathcal{D} = \{(Q_i, A_i, P_i, R_i)\}$, where $P_i$ is a KG reasoning path and $R_i$ is the reasoning trace.
   \STATE {\bfseries Output:} Compositional reasoning model $\theta_{RL}$.
   
   \STATE \COMMENT{\textbf{Stage 1: Supervised Fine-Tuning (SFT)}}
   \STATE Initialize $\theta_{SFT} \leftarrow \theta_{base}$
   \FOR{each $(Q, A, P, R) \in \mathcal{D}_{SFT}$}
       \STATE Fine-tune $\theta_{SFT}$ to minimize $\mathcal{L}_{SFT} = -\log P(A, R | Q; \theta_{SFT})$
       \STATE \COMMENT{Model learns atomic facts and reasoning structure}
   \ENDFOR
   
   \STATE \COMMENT{\textbf{Stage 2: Path-Aligned Reinforcement Learning (RL)}}
   \STATE Initialize $\theta_{RL} \leftarrow \theta_{SFT}$
   \FOR{each question $Q \in \mathcal{D}_{RL}$}
       \STATE Generate $G$ independent outputs $\{O_1, O_2, \dots, O_G\}$ from $\theta_{RL}$
       \FOR{each output $O_g$}
           \STATE Extract reasoning triples $\hat{T}_g$ from $\langle\text{think}\rangle$ block
           \STATE Compute binary reward $R_{bin}$ based on final answer correctness
           \STATE Compute path reward $R_{path}$ by verifying triples in $\hat{T}_g$ against $\mathcal{P}$
           \STATE $R(O_g) = \alpha \cdot R_{bin} + \beta \cdot R_{path}$ \COMMENT{Weighted reward signal}
       \ENDFOR
       \STATE Compute Advantage and update $\theta_{RL}$ using GRPO objective
   \ENDFOR
\end{algorithmic}
\end{algorithm}

\newpage
\section{Model Prompt}
\label{Appendix H}
We employ a structured prompting strategy to enforce the separation of internal reasoning and final answer selection. This structure is critical for the path alignment reward, $R_{path}$, as it enables our reward function to isolate the $think$ block for axiomatic triple extraction. The exact prompt is shown in Fig.~\ref{fig:prompt_template}.

\begin{figure}[H]
\centering
\begin{tcolorbox}[colback=gray!5, colframe=black!75, width=0.95\columnwidth, 
                  boxrule=0.5pt, arc=2pt]
\small
\textbf{System Prompt:} A conversation between user and assistant. The user asks a single-choice Multiple Choice Question, and the assistant solves it using step-by-step reasoning. Please answer the multiple-choice question by selecting only one from option A, option B, option C, or option D. The assistant first thinks through the problem systematically, then provides the explanation and final answer. Use \texttt{<think>...</think>} tags for internal reasoning, then provide the answer in the format --- \texttt{Final Answer: }.

\vspace{0.4em}
\textbf{Task Instructions:} Please provide complete and accurate answers with clear reasoning. The answer must only be a single letter from A, B, C, D.

\vspace{0.4em}
\textbf{Format Enforcement:}

\texttt{<think> [Reasoning Path] </think>}

\texttt{Final Answer: [Letter]}
\end{tcolorbox}
\caption{Prompt template for GRPO training.}
\label{fig:prompt_template}
\end{figure}

%%%%%%%%%%%%%%%%%%%%%%%%%%%%%%%%%%%%%%%%%%%%%%%%%%%%%%%%%%%%%%%%%%%%%%%%%%%%%%%
%%%%%%%%%%%%%%%%%%%%%%%%%%%%%%%%%%%%%%%%%%%%%%%%%%%%%%%%%%%%%%%%%%%%%%%%%%%%%%%

\end{document}